\documentclass[lettersize,journal,twoside]{IEEEtran}
\usepackage{amsmath,amsfonts}
\usepackage{stfloats}
\usepackage{url}
\usepackage{graphicx}
\usepackage{cite}
\usepackage{booktabs}
\usepackage{multirow}
\usepackage{threeparttable}
\usepackage{pifont}
\usepackage{setspace}
\usepackage{mathrsfs}
\usepackage{enumitem}
\usepackage{float}
\usepackage{ulem}
\usepackage{amssymb}
\usepackage{colortbl}

\usepackage{fancyhdr}

\usepackage{hyperref}
\usepackage[switch]{lineno}
\hypersetup{
    colorlinks=true,
    linkcolor=blue,
    filecolor=black,
    urlcolor=black,
    citecolor=purple
}
\usepackage[table,xcdraw,dvipsnames]{xcolor}
\usepackage{scalerel}
\usepackage{tikz}
\usetikzlibrary{svg.path}

\definecolor{orcidlogocol}{HTML}{A6CE39}
\tikzset{
  orcidlogo/.pic={
    \fill[orcidlogocol] svg{M256,128c0,70.7-57.3,128-128,128C57.3,256,0,198.7,0,128C0,57.3,57.3,0,128,0C198.7,0,256,57.3,256,128z};
    \fill[white] svg{M86.3,186.2H70.9V79.1h15.4v48.4V186.2z}
                 svg{M108.9,79.1h41.6c39.6,0,57,28.3,57,53.6c0,27.5-21.5,53.6-56.8,53.6h-41.8V79.1z M124.3,172.4h24.5c34.9,0,42.9-26.5,42.9-39.7c0-21.5-13.7-39.7-43.7-39.7h-23.7V172.4z}
                 svg{M88.7,56.8c0,5.5-4.5,10.1-10.1,10.1c-5.6,0-10.1-4.6-10.1-10.1c0-5.6,4.5-10.1,10.1-10.1C84.2,46.7,88.7,51.3,88.7,56.8z};
  }
}
\newcommand\orcidicon[1]{\href{https://orcid.org/#1}{\mbox{\scalerel*{
\begin{tikzpicture}[yscale=-1,transform shape]
\pic{orcidlogo};
\end{tikzpicture}
}{|}}}}
\newcommand{\settablefont}{\fontsize{7.6}{10.8}\selectfont}

\begin{document}
\normalem
\bstctlcite{IEEEexample:BSTcontrol}

\title{\Huge Integrating Disparity Confidence Estimation into Relative Depth Prior-Guided Unsupervised Stereo Matching}

\author{Chuang-Wei Liu$^{\orcidicon{0000-0003-0260-6236}\,}$, Mingjian Sun$^{\orcidicon{0000-0001-8719-524X}\,}$,~\IEEEmembership{Member,~IEEE}, Cairong Zhao$^{\orcidicon{0000-0001-6745-9674}\,}$, Hanli Wang$^{\orcidicon{0000-0002-9999-4871}\,}$,~\IEEEmembership{Senior Member,~IEEE}, \\
Alexander Dvorkovich$^{\orcidicon{0000-0003-1190-3582}\,}$ and Rui Fan$^{\orcidicon{0000-0003-2593-6596}\,}$,~\IEEEmembership{Senior Member,~IEEE}
\thanks{
Chuang-Wei Liu is with the College of Electronics and Information Engineering, Tongji University, Shanghai 201804, China (e-mail: cwliu@tongji.edu.cn).
}
\thanks{
Mingjian Sun is with the Department of Control Science and Engineering, the Harbin Institute of Technology, Harbin, Heilongjiang, 150001, China, as well as with the Harbin Institute of Technology Suzhou Research Institute, Suzhou, Jiangsu, 215000, China (e-mail: sunmingjian@hit.edu.cn).
}
\thanks{
Cairong Zhao is with the Department of Computer Science and Technology, Tongji University, Shanghai 201804, China (e-mail: zhaocairong@tongji.edu.cn).
}
\thanks{
Hanli Wang is with the College of Electronics and Information Engineering, the School of Computer Science and Technology, and the Key Laboratory of Embedded System and Service Computing (Ministry of Education), Tongji University, Shanghai 201804, China (e-mail: hanliwang@tongji.edu.cn).
}
\thanks{
Alexander Dvorkovich is with the Multimedia Technology and Telecom Department, Telecommunications Center, Moscow Institute of Physics and Technology, 141701, Institutsky Lane, 9, Dolgoprudny, Moscow Region, Russian Federation (e-mail: dvork.alex@gmail.com).
}
\thanks{
Rui Fan is with the College of Electronics and Information Engineering, Shanghai Institute of Intelligent Science and Technology, Shanghai Research Institute for Intelligent Autonomous Systems,  Shanghai Key Laboratory of Intelligent Autonomous Systems, State Key Laboratory of Autonomous Intelligent Unmanned Systems, and Frontiers Science Center for Intelligent Autonomous Systems of the Ministry of Education, Tongji University, Shanghai 201804, China (e-mail: rui.fan@ieee.org).
}
}

\markboth{IEEE Transactions on Circuits and Systems for Video Technology}{Liu \MakeLowercase{\textit{et al.}}: Integrating Disparity Confidence Estimation into Relative Depth Prior-Guided Unsupervised Stereo Matching}

\maketitle

\begin{abstract}
Unsupervised stereo matching has garnered significant attention for its independence from costly disparity annotations. Typical unsupervised methods rely on the multi-view consistency assumption for training networks, which suffer considerably from stereo matching ambiguities, such as repetitive patterns and texture-less regions. A feasible solution lies in transferring 3D geometric knowledge from a relative depth map to the stereo matching networks. However, existing knowledge transfer methods learn depth ranking information from randomly built sparse correspondences, which makes inefficient utilization of 3D geometric knowledge and introduces noise from mistaken disparity estimates. This work proposes a novel unsupervised learning framework to address these challenges, which comprises a plug-and-play disparity confidence estimation algorithm and two depth prior-guided loss functions. Specifically, the local coherence consistency between neighboring disparities and their corresponding relative depths is first checked to obtain disparity confidence. Afterwards, quasi-dense correspondences are built using only confident disparity estimates to facilitate efficient depth ranking learning. Finally, a dual disparity smoothness loss is proposed to boost stereo matching performance at disparity discontinuities. Experimental results demonstrate that our method achieves state-of-the-art stereo matching accuracy on the KITTI Stereo benchmarks among all unsupervised stereo matching methods.
\end{abstract}


\section{Introduction}
\label{sec.intro}

\begin{figure}[t!]
	\begin{center}
		\includegraphics[width=0.45\textwidth]{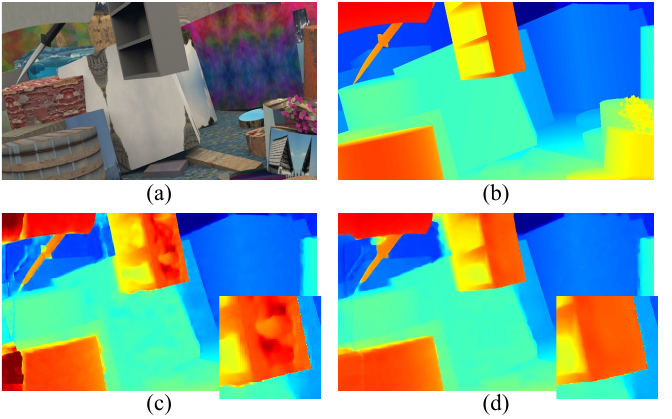}
		\caption{(a) Left image, (b) disparity ground truth, and estimated disparity maps generated by ViTAStereo \cite{liu2024playing} trained using (c) the CDR loss in SC-DepthV3 \cite{sun2023sc} and (d) our proposed LDR loss.}
		\label{fig.intro}
	\end{center}
\end{figure}

\IEEEPARstart{S}{tereo} matching \cite{lu2021resource,zeng2023deep} refers to a process similar to human binocular vision that provides metric depth perception using a rectified stereo vision system \cite{liu2023stereo}. Among all metric depth perception methods, stereo matching has emerged as a prevalent option due to its practicality and cost-effectiveness \cite{fan2018road}. As a result, stereo matching has found wide applications related to geometry 3D reconstruction, including autonomous driving \cite{fan2021learning,fan2020sne}, augmented reality \cite{liu2023low}, image compression \cite{deng2023masic}, remote sensing \cite{li2023whu}, and defect detection \cite{fan2019pothole, fan2021graph}. 

With recent advances in deep learning, researchers have increasingly resorted to convolutional neural networks (CNNs) to address the stereo matching task \cite{lee2021high,chen2023unambiguous,fan2021rethinking}. These CNN-based approaches are capable of learning and aggregating abstract features directly from input stereo images and output disparity maps in an end-to-end fashion \cite{zhang2021farther,li2024inter,okae2021robust,yao2021decomposition}. Witnessing the impressive performance of vision foundation models (VFMs) in deep feature representation and monocular depth estimation, recent supervised studies \cite{bartolomei2025stereo,cheng2025monster} propose to integrate geometric constraints with robust priors from monocular depth VFMs, thereby achieving improved stereo matching accuracy and generalizability. However, a persistent challenge in this research domain lies in the absence of large-scale real-world datasets with disparity ground truth, which significantly restrains the supervised learning of stereo matching \cite{joung2019unsupervised}. Therefore, there is a strong motivation to develop powerful unsupervised stereo matching approaches that are independent of disparity ground truth for learning the best network parameters.

Existing unsupervised stereo matching methods typically rely on the multi-view consistency assumption for network training, such as the commonly adopted photometric loss and geometry consistency loss that enforce the local block or pixel-wise similarity between correspondences \cite{liu2023stereo}. This assumption provides effective constraints for learning 3D geometric knowledge in general scenes, while exhibiting poor robustness in areas with stereo ambiguities, \textit{e.g.}, repetitive patterns and texture-less regions. The matching clues for disparity estimation at these ambiguous areas are insufficient to guide CNNs toward the correct optimization direction, thereby hindering effective parameter convergence \cite{yang2018segstereo}. Inspired by the aforementioned VFM-based yet supervised stereo matching methods, this work aims to address stereo matching ambiguities by integrating robust geometric priors derived from monocular depth VFMs into existing unsupervised learning frameworks.

To address the limitations of the multi-view consistency assumption, recent studies \cite{yang2018segstereo,ren2020suw,sun2023sc} aim at transferring 3D geometric knowledge from depth priors or multi-model data into metric depth perception CNNs. For instance, SC-DepthV3 \cite{sun2023sc} first separates foreground and background areas based on a relative depth map and then establishes correspondence for each foreground pixel with another randomly selected background pixel. Afterwards, a confident depth ranking (CDR) loss is introduced to enforce divergent estimates between the built correspondences. Nevertheless, this knowledge transfer method still suffers from the following limitations:
\begin{itemize}
    \item The randomly generated correspondences introduce substantial noise to the loss functions, especially those built upon inaccurate disparity estimates. 
    \item The immensely sparse correspondences lead to inefficient exploitation of the 3D geometric knowledge.
    \item The simplistic disparity divergence strategy fails to provide precise guidance for disparity estimation. 
\end{itemize}

Motivated by these issues, this work aims at enhancing the efficiency and accuracy of the knowledge transfer process by selectively building quasi-dense correspondences with only reliable disparity estimates. We first propose an unsupervised disparity confidence estimation algorithm, named \textbf{Disparity-Depth Consistency Voting (DDCV)}. In DDCV, each disparity undergoes a voting process within its neighborhood system, with the average voting score representing its confidence. Each vote is either $0$ or $1$, denoting a negative or positive vote, respectively, and is derived by checking the local coherence consistency between the disparity pairs and their corresponding relative depth pairs. Afterwards, we introduce a \textbf{local depth ranking (LDR)} loss function to learn the depth rankings by building quasi-dense correspondences with reliable disparities as suggested by DDCV. In general, the quasi-dense correspondences enhance the knowledge transfer efficiency, and the reliable disparities suggested by DDCV effectively suppress noise introduced by incorrect estimates. Moreover, LDR loss penalizes only those disparity correspondences that exhibit inconsistency with their corresponding relative depth rankings. This strategy enables our LDR loss to deliver more precise guidance for network parameter convergence. Finally, we propose a novel \textbf{dual disparity smoothness (DDS)} loss function to reinforce the consistency between the estimated disparities and relative depth priors in terms of both local smoothness and depth discontinuities. A comparison between the CDR loss in SC-DepthV3 and our proposed LDR loss is presented in Fig. \ref{fig.intro}. It can be observed that our proposed LDR loss leads to improved disparity smoothness and more distinct boundaries at disparity discontinuities. We conclude the contributions of this study as follows: 
\begin{itemize}
    \item We discover the significance of disparity confidence in effectively transferring 3D geometric knowledge from relative depth priors to stereo matching networks.
    \item We propose DDCV, a real-time and plug-and-play algorithm that estimates disparity confidence from any disparity map and its corresponding relative depth map.
    \item We introduce two relative depth prior-guided loss functions for unsupervised stereo matching, leading to \uline{\textbf{state-of-the-art (SoTA) stereo matching accuracy}} on the KITTI Stereo benchmark among all existing unsupervised methods.
\end{itemize}

The remainder of this article is structured as follows: related works, including unsupervised stereo matching and disparity confidence estimation methods, are presented in Sect. \ref{sec.related}. Sect. \ref{sec.method} details our proposed DDCV algorithm, LDR and DDS loss functions. Comprehensive ablation studies and comparative experiments are presented in Sect. \ref{sec.exp}. Finally, in Sect. \ref{sec.conc}, we summarize the results and provide recommendations for future work. Our supplementary material and code are publicly available at \url{https://mias.group/Un-ViTAStereo}.

\section{Related Work}
\label{sec.related}

\subsection{Unsupervised Stereo Matching}

Extensive efforts have been devoted to unsupervised stereo matching to eliminate the dependence on costly disparity ground truth. Under the commonly adopted multi-view consistency assumption, these methods typically warp the right image and disparity map into the left view using the estimated left disparity map \cite{uddin2022unsupervised,yang2024self}. Afterwards, a hybrid loss, such as a combination of photometric loss and geometry consistency loss, is employed to enforce similarity between the reconstructed left views and the original ones \cite{zhong2017self}. Some works attempt to incorporate occlusion inference into the network training process to enforce the structural understanding \cite{li2018occlusion,li2021unsupervised}. For instance, Co-Teaching \cite{wang2021co} employs a co-teaching framework, where two networks with different initializations interactively teach each other to improve occlusion handling. Other studies enhance stereo matching accuracy by jointly learning stereo matching with optical flow estimation and ego-motion, without relying on any post-processing. For example, Flow2Stereo \cite{liu2020flow2stereo} trains one single network to estimate both flow and disparity, thereby fully exploiting the inherent geometric constraints from stereoscopic videos. In contrast to existing methods, our work novelly proposes to transfer the 3D geometric knowledge from relative depth priors into a stereo matching network. Notably, our method requires no modifications to the network architecture and optimization of the training data, yet achieves SoTA stereo matching accuracy among all unsupervised methods on public benchmarks.

Extensive efforts have been devoted to unsupervised stereo matching to eliminate the dependence on costly disparity ground truth. Under the commonly adopted multi-view consistency assumption, these methods typically warp the right image and disparity map into the left view using the estimated left disparity map \cite{yang2024self}. Afterwards, a hybrid loss, such as a combination of photometric loss and geometry consistency loss, is employed to enforce similarity between the reconstructed left views and the original ones \cite{zhong2017self}. Some works attempt to incorporate occlusion inference into the network training process to enforce the structural understanding \cite{li2018occlusion,li2021unsupervised}. For instance, Co-Teaching \cite{wang2021co} employs a co-teaching framework, where two networks with different initializations interactively teach each other to improve occlusion handling. Other studies enhance stereo matching accuracy by jointly learning stereo matching with optical flow estimation and ego-motion, without relying on any post-processing. For example, Flow2Stereo \cite{liu2020flow2stereo} trains one single network to estimate both flow and disparity, thereby fully exploiting the inherent geometric constraints from stereoscopic videos. In contrast to existing methods, our work novelly proposes to transfer the 3D geometric knowledge from relative depth priors into a stereo matching network. Notably, our method requires no modifications to the network architecture and optimization of the training data, yet achieves SoTA stereo matching accuracy among all unsupervised methods on public benchmarks.

\subsection{Knowledge Transfer}

Aiming to learn the depth rankings from a relative depth map, our proposed loss functions are thus also related to knowledge transfer methods. For instance, SUW-learn \cite{ren2020suw} learns 3D geometric knowledge from a pre-calculated semantic segmentation map under the assumption that pixels with different semantic labels exhibit significantly different depth rankings. Specifically, correspondences are established between each pixel and another randomly selected pixel. Afterwards, a loss function is introduced to encourage similar disparity estimates for correspondences with the same semantic label, and divergent estimates for those with different semantic labels. SC-DepthV3 \cite{sun2023sc} builds upon this strategy by learning more fine-grained depth rankings from a pre-calculated relative depth map, while making minor improvements in correspondence density and the disparity divergence strategy. By contrast, our proposed LDR loss significantly improves knowledge transfer accuracy and efficiency by building quasi-dense correspondences with only confident disparity estimates.

\subsection{Disparity Confidence Estimation}

Disparity confidence estimation methods are either explicit programming-based or data-driven \cite{poggi2021confidence}. The former ones mainly use cost volumes as the source of confidence information, such as the difference or ratio between two cost minima, as in the peak ratio naive \cite{hu2012quantitative} and maximum margin \cite{poggi2017quantitative}. More recent methods introduce various geometric assumptions to estimate confidence directly from the disparity map. For instance, the disparity agreement \cite{poggi2016learningo1,poggi2020learning} assigns higher confidence to disparities that exhibit greater local coherence, and the uniqueness constraint \cite{poggi2020self} requires a reliable disparity to establish exclusive correspondence in the reference view. Early data-driven methods estimate disparity confidence from small patches of the disparity map, such as CCNN \cite{poggi2016learning}, which introduces CNN into this research domain for the first time. The local-global confidence network \cite{tosi2018beyond} leverages a U-Net architecture \cite{ronneberger2015u} and estimates disparity confidence from both the image and disparity map. SEDNet \cite{chen2023learning} proposes a differentiable soft-histogramming mechanism to model the error between intermediate multi-scale disparity maps for confidence estimation. 

However, most data-driven methods are supervised and rely heavily on disparity ground truth. In contrast, unsupervised learning has been barely investigated in this research domain. Existing unsupervised methods generally use a combination of explicit-based methods to generate pseudo-supervision signals for random forest \cite{tosi2017learning} classifiers or CNNs \cite{park2018learning,poggi2020self}. The study presented in \cite{mostegel2016using} proposes to generate positive and negative training samples by analyzing the consistency between disparity maps of the same static scene. Despite these efforts, a substantial performance gap remains between unsupervised and supervised CNNs. In contrast, our proposed DDCV is an explicit programming-based and, to the best of our knowledge, \uline{\textbf{the first exploration}} of relative depth prior-guided disparity confidence estimation algorithm. Without requiring any learnable parameters, DDCV achieves comparable accuracy with recent SoTA supervised disparity confidence estimation methods. Therefore, we regard DDCV as a strong baseline for future powerful relative depth-guided disparity confidence estimation methods. 

\begin{figure*}[t!]
	\begin{center}
		\includegraphics[width=0.99\textwidth]{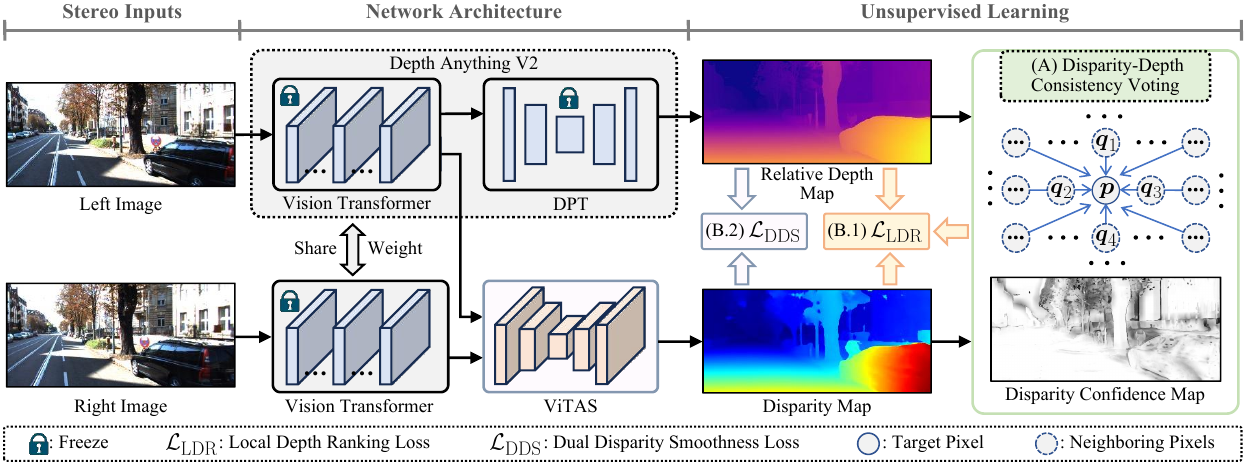}
		\caption{An illustration of the model architecture and training framework in this work. DPT represents the dense prediction Transformer \cite{chen2023vision}, serving as the decoder of a pre-trained Depth Anything V2 model. Our ViTAStereo shares the feature encoder with the Depth Anything V2 model \cite{yang2024depthv2}. The parameters of the feature encoder are frozen.}
		\label{fig.framework}
	\end{center}
\end{figure*}

\section{Methodology}
\label{sec.method}

Recent advances in VFMs \cite{kirillov2023segment,yang2024depth} have further empowered knowledge transfer-based unsupervised stereo matching methods by providing robust relative depth priors. Additionally, our previous work, ViT Adapter for Stereo (ViTAS) \cite{liu2024playing}, successfully adapts the VFM features to enhance the performance of a stereo matching network. Therefore, in this work, we incorporate ViTAS and the pre-trained vision Transformer encoder of the Depth Anything V2 model \cite{yang2024depthv2} to generate ViTAStereo \cite{liu2024playing}, which is able to generate the disparity map $\boldsymbol{D}$ and its corresponding relative depth map $\widetilde{\boldsymbol{D}}$ at a low computational cost, as visualized in Fig. \ref{fig.framework}. The yielded disparity and relative depth maps are subsequently used to initialize our unsupervised learning framework. Therefore, this work can also be regarded as the unsupervised learning version of ViTAStereo. Notably, although relative depth maps are required in the network training process, the disparity inference process takes as input only the stereo image pairs, making our unsupervised learning framework compatible with any other existing stereo matching networks. The proposed disparity confidence estimation algorithm, DDCV, is detailed in Sect. \ref{sec.DDCV}. Afterwards, all loss functions adopted in this work are introduced in Sect. \ref{sec.loss}.

\subsection{Disparity-Depth Consistency Voting}
\label{sec.DDCV}

Building upon the concept of local coherence constraint \cite{roy1999stereo}, Markov Random Field (MRF)-based stereo matching methods \cite{fan2018road,liu2025these} have remained both prevalent and competitive in this research domain, regardless of the era—before or after the emergence of CNNs. These methods suggest that a crucial factor in accurate disparity estimation lies in discovering disparities that exhibit local coherence with their neighboring disparities. Notably, the relative depths derived from the Depth Anything V2 model provide sufficient local coherence priors between neighboring pixels. Drawing inspiration from the success of these MRF-based methods, we introduce DDCV, a disparity confidence estimation algorithm that models the proportion of neighboring disparity pairs that exhibit local coherence consistency with their corresponding relative depth pairs. In DDCV, the confidence of each disparity is calculated through a dense voting process within its neighborhood system using two proposed local coherence-based voting functions: the ranking consistency voting function $\mathbb{F}_{\text{RC}}$ and the variation consistency voting function $\mathbb{F}_{\text{VC}}$. Each vote is either 1 or 0, representing a positive or negative vote, respectively. Finally, the confidence of a disparity is derived by averaging all votes from its neighborhood system.

Given an imperfectly estimated disparity map $\boldsymbol{D}$ and its corresponding relative depth map $\widetilde{\boldsymbol{D}}$. A neighborhood system $\mathcal{N}_{p}=\{\boldsymbol{q}_{1},\ldots,\boldsymbol{q}_{m}\}$ comprises a collection of 2D pixels centered at a target pixel $\boldsymbol{p}$. $\mathbb{F}_{\text{RC}}$ imposes a local coherence constraint on the depth ranking between each disparity pair and its corresponding relative depth pair within $\mathcal{N}_{p}$ as follows:
\begin{equation}
\mathbb{F}_{\text{RC}}(\boldsymbol{p},\boldsymbol{q}) = \Theta(\triangle\widetilde{\boldsymbol{D}}_{pq}\times\triangle\boldsymbol{D}_{pq}),
\end{equation}
where $\times$ operation denotes the product of two scalar values, $\triangle\widetilde{\boldsymbol{D}}_{pq} = \widetilde{\boldsymbol{D}}(\boldsymbol{p}) - \widetilde{\boldsymbol{D}}(\boldsymbol{q})$ and $\triangle\boldsymbol{D}_{pq} = \boldsymbol{D}(\boldsymbol{p}) - \boldsymbol{D}(\boldsymbol{q})$ represent the disparity variation and depth variation, respectively, and
\begin{equation}
\Theta(x) = 
    \begin{cases} 
       1,   &  x \geqslant 0, \\
       0,   &  \text{otherwise,}
    \end{cases}
\end{equation}
is a step function to decide the depth ranking consistency vote. In general, a positive depth ranking consistency vote requires the depth ranking between a pair of neighboring pixels to remain consistent between $\boldsymbol{D}$ and $\widetilde{\boldsymbol{D}}$.

\begin{figure}[t!]
	\begin{center}
		\includegraphics[width=0.49\textwidth]{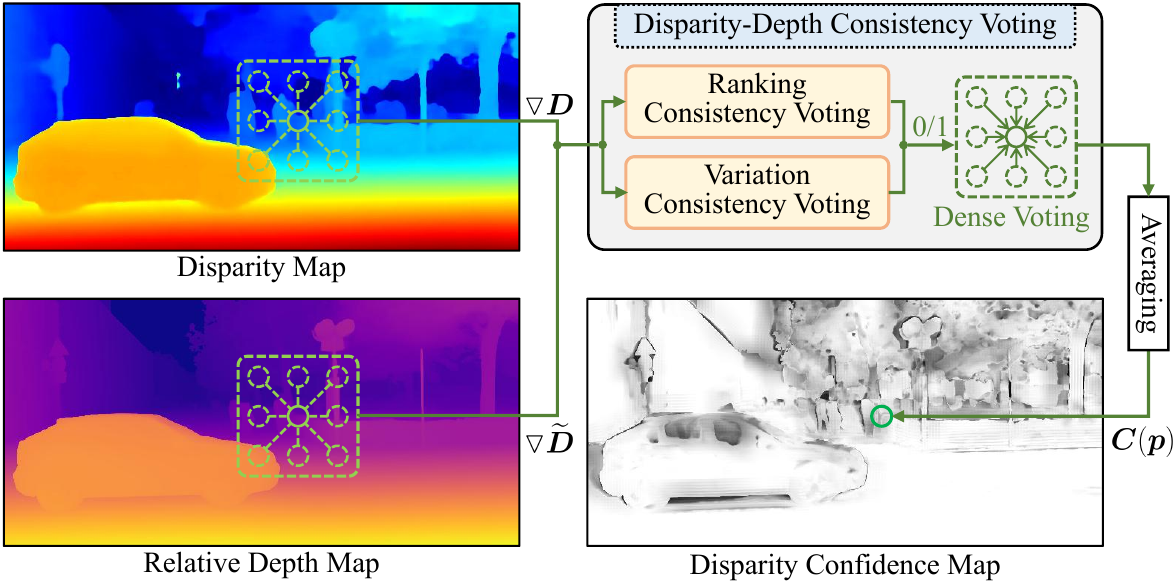}
		\caption{An illustration of our proposed DDCV.}
		\label{fig.DDCV}
	\end{center}
\end{figure}

The variation consistency voting function $\mathbb{F}_{\text{VC}}$ requires mild disparity variations in regions with stable relative depth, and significant disparity variations at relative depth discontinuities, as follows:
\begin{equation}
\begin{split}
\mathbb{F}_{\text{VC}}(\boldsymbol{p},\boldsymbol{q}) = &\Theta(\Theta(|\triangle\widetilde{\boldsymbol{D}}_{pq}|-\gamma\times\sigma)\times(1-|\triangle\boldsymbol{D}_{pq}|)) \\
\times &\Theta(\Theta(\gamma-|\triangle\widetilde{\boldsymbol{D}}_{pq}|) \times (|\triangle\boldsymbol{D}_{pq}| - \sigma)),
\end{split}
\label{eq.alpha}
\end{equation}
where
\begin{equation}
\gamma=\frac{\sum_{\boldsymbol{p} \in \widetilde{\boldsymbol{D}}}\sum_{\boldsymbol{q}\in\mathcal{N}_{p}} |\triangle\widetilde{\boldsymbol{D}}_{pq}|}{\sum_{\boldsymbol{p} \in \boldsymbol{D}}\sum_{\boldsymbol{q}\in\mathcal{N}_{p}} |\triangle\boldsymbol{D}_{pq}|},
\end{equation} 
denotes a rough global scale factor, $\sigma \in (1,+\infty)$ is a hyper-parameter to loosen the variation consistency constraint, thereby improving the robustness of $\mathbb{F}_{\text{VC}}$ in areas with a local scale factor that differs significantly from $\gamma$. In $\mathbb{F}_{\text{VC}}$, $\Theta(\cdot)$ is also used to identify regions with stable or discontinuous relative depths. Afterwards, the final vote from $\boldsymbol{q}$ to $\boldsymbol{p}$ is calculated as follows:
\begin{equation}
v_{{pq}} = \mathbb{F}_{\text{RC}}(\boldsymbol{p},\boldsymbol{q}) \times \mathbb{F}_{\text{VC}}(\boldsymbol{p},\boldsymbol{q}), 
\end{equation} 
where any local coherence inconsistency results in a negative vote. Finally, the confidence of disparity $\boldsymbol{D}(\boldsymbol{p})$ is computed by averaging all votes from its neighborhood system:
\begin{equation}
\mathcal{C}(\boldsymbol{p}) = \frac{\sum_{\boldsymbol{q}\in\mathcal{N}_{p}} v_{pq}}{m}.
\label{eq.confidence}
\end{equation}
The confidence of all disparities collectively forms the final output confidence map $\mathcal{C}$ of DDCV. More details are illustrated in Fig. \ref{fig.DDCV}. Moreover, the relative depths generated by the Depth Anything V2 model present larger values at objects closer to the camera. Therefore, the inverse proportional relationship between the disparity and metric depth is omitted in this work.

\subsection{Unsupervised Loss Functions}
\label{sec.loss}

The hybrid loss function used in this work comprises four items as follows:
\begin{equation}
\mathcal{L} = \mathcal{L}_{\text{P}} + \lambda_{1} \times \mathcal{L}_{\text{LRC}} + \lambda_{2} \times \mathcal{L}_{\text{LDR}} + \lambda_{3} \times \mathcal{L}_{\text{DDS}},
\label{eq.loss}
\end{equation}
where $\lambda_{1}$, $\lambda_{2}$ and $\lambda_{3}$ are three hyper-parameters to weight the contributions of the loss terms, and $\mathcal{L}_{\text{P}}$ and $\mathcal{L}_{\text{LRC}}$, denoting the photometric loss and left-right consistency loss, respectively, are two widely adopted loss functions in unsupervised stereo matching methods \cite{zhong2017self,wang2021pvstereo}. The photometric loss $\mathcal{L}_{\text{P}}$ is defined as follows:
\begin{equation}
\begin{split}
\mathcal{L}_{\text{P}} & = \frac{1}{N} \sum_{\boldsymbol{p} \in \boldsymbol{I}^{L}} (0.85 \times \frac{1-\text{SSIM}(\boldsymbol{I}^{L}(\boldsymbol{p}),\hat{\boldsymbol{I}}^{R}(\boldsymbol{p}))}{2}\\
& + 0.15 \times \|\boldsymbol{I}^{L}(\boldsymbol{p})-\hat{\boldsymbol{I}}^{R}(\boldsymbol{p})\|_{1}),
\end{split}
\end{equation}
where $\boldsymbol{I}^{L}$ represents the left stereo image, $\hat{\boldsymbol{I}}^{R}$ is generated by warping the original right stereo image $\boldsymbol{I}^{R}$ into the left view using the estimated disparity map $\boldsymbol{D}$, $N$ denotes the number of pixels within $\boldsymbol{I}^{L}$, and SSIM, denoting the structural similarity index measure \cite{wang2004image}, is used to measure image similarity. The left-right consistency loss $\mathcal{L}_{LRC}$ is defined as follows:
\begin{equation}
\mathcal{L}_{\text{LRC}} = \frac{1}{N} \sum_{\boldsymbol{p} \in \boldsymbol{I}^{L}} \frac{|\boldsymbol{D}(\boldsymbol{p})-\hat{\boldsymbol{D}}^{R}(\boldsymbol{p})|}{\boldsymbol{D}(\boldsymbol{p})+\hat{\boldsymbol{D}}^{R}(\boldsymbol{p})},
\end{equation}
where $\hat{\boldsymbol{D}}^{R}$ is derived by warping the right disparity map $\boldsymbol{D}^{R}$ into the left view using $\boldsymbol{D}$. Our proposed LDR loss $\mathcal{L}_{\text{LDR}}$ and DDS loss $\mathcal{L}_{\text{DDS}}$ are detailed in the subsequent subsections. 

\subsubsection{Local Depth Ranking Loss}
\label{sec.OC}

We develop our proposed LDR loss $\mathcal{L}_{\text{LDR}}$ based on the disparity continuity hypothesis that disparities change gradually across continuous regions \cite{liu2025these}. For each disparity, its neighboring disparities converge toward it from similar values, and the size relationships between them are determined by their corresponding relative depth rankings. Given that most disparities do not correspond to a local depth extremum, their neighboring disparities are distributed on both the greater and smaller sides, allowing these disparities to be precisely confined within a closed interval. As a result, erroneous disparity estimates, particularly those arising from stereo matching ambiguities, can be effectively rectified using these theoretical disparity ranges.

With the disparity confidence map initialized by DDCV, pixels with high confidence disparities are first selected as reference points, serving as local depth anchors for their neighboring pixels. Afterwards, correspondences are established for each disparity with its neighboring reference points. Finally, our proposed $\mathcal{L}_{\text{LDR}}$ learns the 3D geometric knowledge from the relative depth priors by penalizing the disparity correspondences that go against their corresponding relative depth rankings.

Specifically, for each pixel $\boldsymbol{p}$, a top-$k$ operation first identifies a set of $k$ reference points $\mathcal{R}_{p}=\{\boldsymbol{r}_{1},\cdots,\boldsymbol{r}_{k}\}$ with the highest disparity confidence within a dilated sampling window centred at $\boldsymbol{p}$. The usage of a dilated sampling window helps prevent spatially clustered reference points compared with a general window, thereby ensuring a more precise value range for $\boldsymbol{D}(\boldsymbol{p})$. Afterwards, we compute $\mathcal{L}_{\text{LDR}}$ between each disparity correspondence as follows:
\begin{equation}
\phi(\boldsymbol{p}) = \frac{\sum_{\boldsymbol{r} \in \mathcal{R}_{p}}|1-\Theta(\boldsymbol{p},\boldsymbol{r})| \times \omega(\triangle\widetilde{\boldsymbol{D}}_{pr}) \times \nu(\triangle\boldsymbol{D}_{pr})}{\sum_{\boldsymbol{r} \in \mathcal{R}_{p}}|1-\Theta(\boldsymbol{p},\boldsymbol{r})| \times \omega(\triangle\widetilde{\boldsymbol{D}}_{pr})},
\end{equation}
where the $|1-\Theta(\boldsymbol{p},\boldsymbol{r})|$ operation identifies the disparity correspondences that exhibit inconsistent metric depth rankings with their corresponding relative depth rankings,
\begin{equation}
\omega(\triangle\widetilde{\boldsymbol{D}}_{pr}) = \frac{|\triangle\widetilde{\boldsymbol{D}}_{pr}|}{1+|\triangle\widetilde{\boldsymbol{D}}_{pr}|},
\end{equation}
assigns a greater weight to pixel correspondences with substantially different relative depths, and
\begin{equation}
\nu(\triangle\boldsymbol{D}_{pr}) = \text{log}(1+|\triangle\boldsymbol{D}_{pr}|),
\end{equation} 
requires the disparity estimates to reside within the disparity value ranges constrained by the relative depth rankings. Finally, our proposed $\mathcal{L}_{\text{LDR}}$ is defined as follows:
\begin{equation}
\mathcal{L}_{\text{LDR}} = \frac{1}{N}\sum_{p \in \boldsymbol{I}^{L}} \phi(\boldsymbol{p}).
\end{equation}
By building quasi-dense correspondences only with neighboring reference points associated with reliable disparities, our proposed $\mathcal{L}_{\text{LDR}}$ enables full and accurate utilization of relative depth priors in the unsupervised learning of stereo matching.

\begin{figure}[t!]
	\begin{center}
		\includegraphics[width=0.49\textwidth]{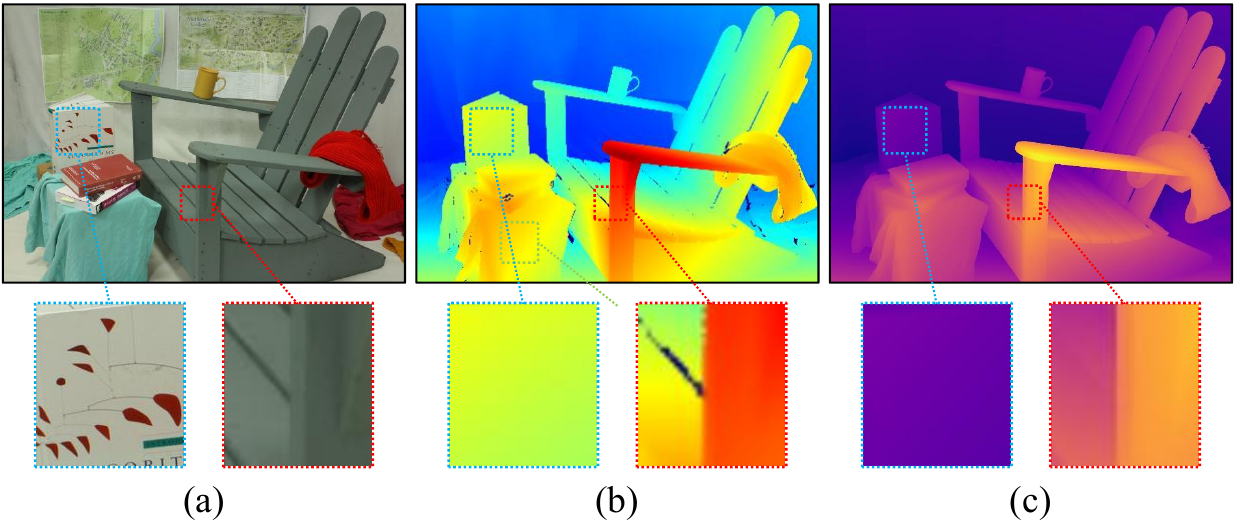}
		\caption{Illustrations of inconsistency between image intensities and disparities in the (a) left image, (b) ground truth disparity map, and (c) relative depth map. The blue boxes highlight regions with discontinuous pixel intensities but smooth depth variations, while the red boxes indicate regions with smooth image intensities but depth discontinuities.}
		\label{fig.smooth}
	\end{center}
\end{figure}

\subsubsection{Dual Disparity Smoothness Loss}
\label{sec.DDS}

The conventional image-based disparity smoothness loss \cite{ren2020suw} enforces local smoothness in disparity estimations by applying an $\mathop{L}_{1}$ penalty on the disparity gradients $\partial \boldsymbol{D}$. Given that disparity discontinuities tend to occur at places where image pixel intensity changes greatly, the disparity gradients are weighted with an edge-aware term using the image gradients $\partial \boldsymbol{I}^{L}$. The general formulation of the image-based disparity smoothness loss is expressed as follows:
\begin{equation}
\mathcal{L}_{\text{DS}}^{I} = \frac{1}{N} \sum_{\boldsymbol{p} \in \boldsymbol{I}^{L}}|\partial\boldsymbol{D}(\boldsymbol{p})|e^{-\|\partial\boldsymbol{I}^{L}(\boldsymbol{p})\|}.
\end{equation}
In general, $\mathcal{L}_{\text{DS}}^{I}$ assumes disparities are locally smooth except in areas with discontinuous image pixel intensities. However, such a simplistic assumption generates incorrect guidance under two common situations: (1) regions with complicated patterns that produce discontinuous image pixel intensities unrelated to depth discontinuities, and (2) depth discontinuities between foreground and background objects that share similar image intensities. 

In contrast, monocular depth estimation networks, such as the Depth Anything V2 model, are able to extract and aggregate global texture information, which allows them to overlook the image patterns and accurately identify boundaries between foreground and background objects that have similar image intensities. The generated relative depth map provides a more faithful representation of metric depth smoothness and discontinuities than the original image, as illustrated in Fig. \ref{fig.smooth}. Consequently, we first propose to replace the image gradients with the relative depth gradients, resulting in a relative depth-based disparity smoothness loss as follows:
\begin{equation}
\mathcal{L}_{\text{DS}}^{\widetilde{D}} = \frac{1}{N} \sum_{\boldsymbol{p} \in \boldsymbol{I}^{L}}|\partial\boldsymbol{D}(\boldsymbol{p})|e^{-\|\partial\widetilde{\boldsymbol{D}}(\boldsymbol{p})\|}.
\end{equation}

Moreover, leveraging the additional depth discontinuity information provided by the relative depths, we further enhance the relative depth-based disparity smoothness loss with a dual term, which gives a large penalty at areas with discontinuous relative depths but smooth disparities. Finally, our proposed $\mathcal{L}_{\text{DDS}}$ is defined as follows:
\begin{equation}
\mathcal{L}_{\text{DDS}} = \frac{1}{N} \sum_{\boldsymbol{p} \in \boldsymbol{I}^{L}}|\partial\boldsymbol{D}(\boldsymbol{p})|e^{-\|\partial\widetilde{\boldsymbol{D}}(\boldsymbol{p})\|} + |\partial\widetilde{\boldsymbol{D}}(\boldsymbol{p})|e^{-\|\partial\boldsymbol{D}(\boldsymbol{p})\|}.
\end{equation}
Compared with the conventional $\mathcal{L}_{\text{DS}}^{I}$, $\mathcal{L}_{\text{DDS}}$ enhances the consistency of both local smoothness and local discontinuity between the estimated disparities and relative depth priors, thereby resulting in better stereo matching performance at depth discontinuities. 

\begin{figure}[t!]
	\begin{center}
		\includegraphics[width=0.4\textwidth]{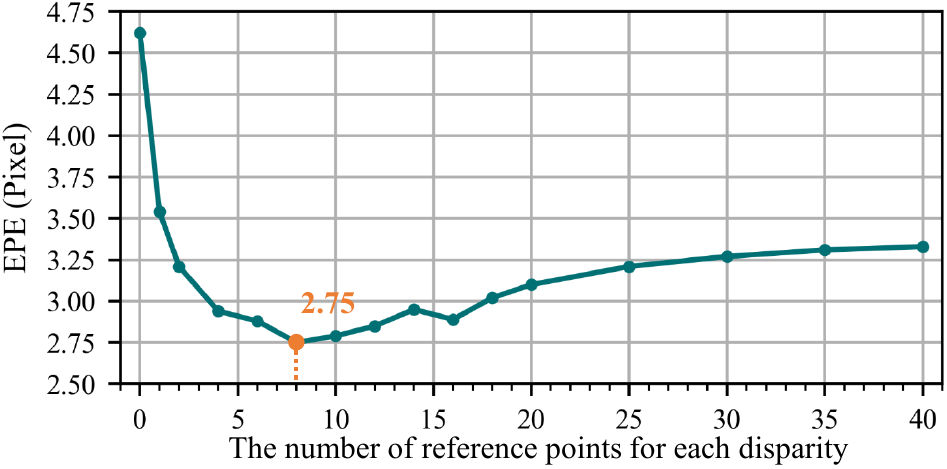}
		\caption{Ablation study on the top-$k$ operation in $\mathcal{L}_{\text{LDR}}$ w.r.t. different $k$.}
		\label{fig.topk}
	\end{center}
\end{figure}

\section{Experiments}
\label{sec.exp}

\subsection{Datasets and Implementation Details}

Five public stereo matching datasets are utilized in our experiments for model training and performance evaluation, including two synthetic large-scale datasets: 
\begin{enumerate}
    \item \textbf{SceneFlow} \cite{mayer2016large} consists of a training set (containing 35,454 stereo image pairs) and a test set (often known as the Flying 3D test set, containing 4,370 stereo image pairs) with image resolution of $960 \times 540$ pixels. 
    \item \textbf{Virtual KITTI} \cite{cabon2020virtual} contains 21,260 stereo image pairs {(resolution: $1,242 \times 375$ pixels)}, generated from five different virtual worlds (created using the Unity game engine and a real-to-virtual cloning method) in urban settings under different imaging and weather conditions,
\end{enumerate}
and three real-world yet small-scale datasets:
\begin{enumerate}
    \item \textbf{KITTI Stereo} contains two subsets: 2012 \cite{geiger2012we} and 2015 \cite{menze2015object}, with 192 and 200 training pairs, respectively, and 194 and 200 test pairs, respectively. The image resolution is around $1,240 \times 370$ pixels. 
    \item \textbf{Middlebury} \cite{scharstein2014high} contains 14 pairs of indoor stereo images and is widely recognized for its complicated 3D geometry structures. 
    \item \textbf{ETH3D} \cite{schops2017multi} contains 27 pairs of stereo grayscale images of both indoor and outdoor scenes (resolution: around $930 \times 490$ pixels).
\end{enumerate}

We first conduct ablation studies on the SceneFlow dataset to validate the effectiveness of each proposed component. Subsequently, ViTAStereo is trained on all five datasets and evaluated on the test sets of the KITTI Stereo datasets. Furthermore, we perform comprehensive comparisons between our proposed DDCV and other SoTA disparity confidence estimation methods on the three real-world datasets under various disparity input settings. We select eight disparities with the highest confidences as reference points in $\mathcal{L}_{\text{LDR}}$ (top-\textit{k} with $\textit{k}=8$) as suggested by our ablation study. We also set $\alpha = 2$ in \eqref{eq.alpha}, $\lambda_1 = \lambda_3 = 0.1$ and $\lambda_2 = 0.1$ in \eqref{eq.loss}, and use an $11 \times 11$ sampling kernel in both DDCV and $\mathcal{L}_{\text{LDR}}$. These parameters are intuitively chosen without extensive hyperparameter tuning. All experiments are conducted on an NVIDIA RTX 4090 GPU. During model training, we randomly crop images to $320 \times 720$ pixels and apply conventional data augmentation techniques, including random changes in image color, random rescaling, and random erasing. 

\subsection{Evaluation Metrics}

Three commonly adopted metrics are computed to quantify stereo matching accuracy: \textbf{End-point error (EPE)} measures the average disparity estimation error, \textbf{Percentage of error pixels (PEP)} indicates the percentage of incorrect disparities with respect to a tolerance of $\delta$ pixels, and \textbf{D1} counts the percentage of disparities for which the estimation error exceeds both three pixels and 5\% of the ground-truth disparity.

Additionally, we generate the density-EPE receiver operating characteristic (ROC) curves and then use the \textbf{area under the curve (AUC)} to evaluate the disparity confidence estimation accuracy following previous work \cite{chen2023learning}. Specifically, the ROC curves in our case measure EPE by successively incorporating disparities in descending order of their confidence. We also introduce the optimal AUC to count the area under the optimal ROC curve that incorporates disparities in ascending order of their EPE values. Moreover, \textbf{time/MP} that measures the runtime in megapixels is used to evaluate the disparity confidence estimation efficiency. For all metrics used in our experiments, lower values indicate better performance. 

\begin{table}[t!]
    \settablefont
    \caption{Ablation studies on the Sceneflow test dataset. The best results are shown in bold type.}
    \centering
    \label{tab.abla}
    \setlength{\tabcolsep}{1.1mm}{
	\begin{tabular}
		{ccccccc}
		\toprule[1.5pt]
		\multicolumn{2}{c}{Depth Ranking} & \multicolumn{3}{c}{Disparity Smoothness} & \multirow{2}{*}{EPE (pixel)} & \multirow{2}{*}{D1 (\%)} \\ 
		\cmidrule(r){1-2}
		\cmidrule(r){3-5}
		\specialrule{0em}{-1pt}{-1pt}
		\makebox[0.05\textwidth][c]{$\mathcal{L}_{\text{CDR}}$} & \makebox[0.05\textwidth][c]{$\mathcal{L}_{\text{LDR}}$} & \makebox[0.05\textwidth][c]{$\mathcal{L}_{\text{DS}}^{I}$} & \makebox[0.05\textwidth][c]{$\mathcal{L}_{\text{DS}}^{\widetilde{D}}$} & \makebox[0.05\textwidth][c]{$\mathcal{L}_{\text{DDS}}$}\\ 
		\midrule[1pt]
            & & & & & 4.91 & 16.6 \\
            \ding{52} & & & & & 3.46 & 13.7 \\
            & {\ding{52}} & & & & 3.09 & 12.4 \\
            \ding{52} & \ding{52} & & & & 3.08 & 12.2 \\
            \midrule
            & \ding{52} & \ding{52} & & & 2.88 & 11.5 \\
            & \ding{52} & \ding{52} & \ding{52} & & 2.80 & 11.0 \\
            & \ding{52} & & \ding{52} & & 2.81 & 11.0 \\
            & \ding{52} & & & \ding{52} & \textbf{2.75} & \textbf{10.7} \\
            \bottomrule [1.5pt]   	
	\end{tabular}}
\end{table}

\begin{figure*}[t!]
	\begin{center}
		\includegraphics[width=0.99\textwidth]{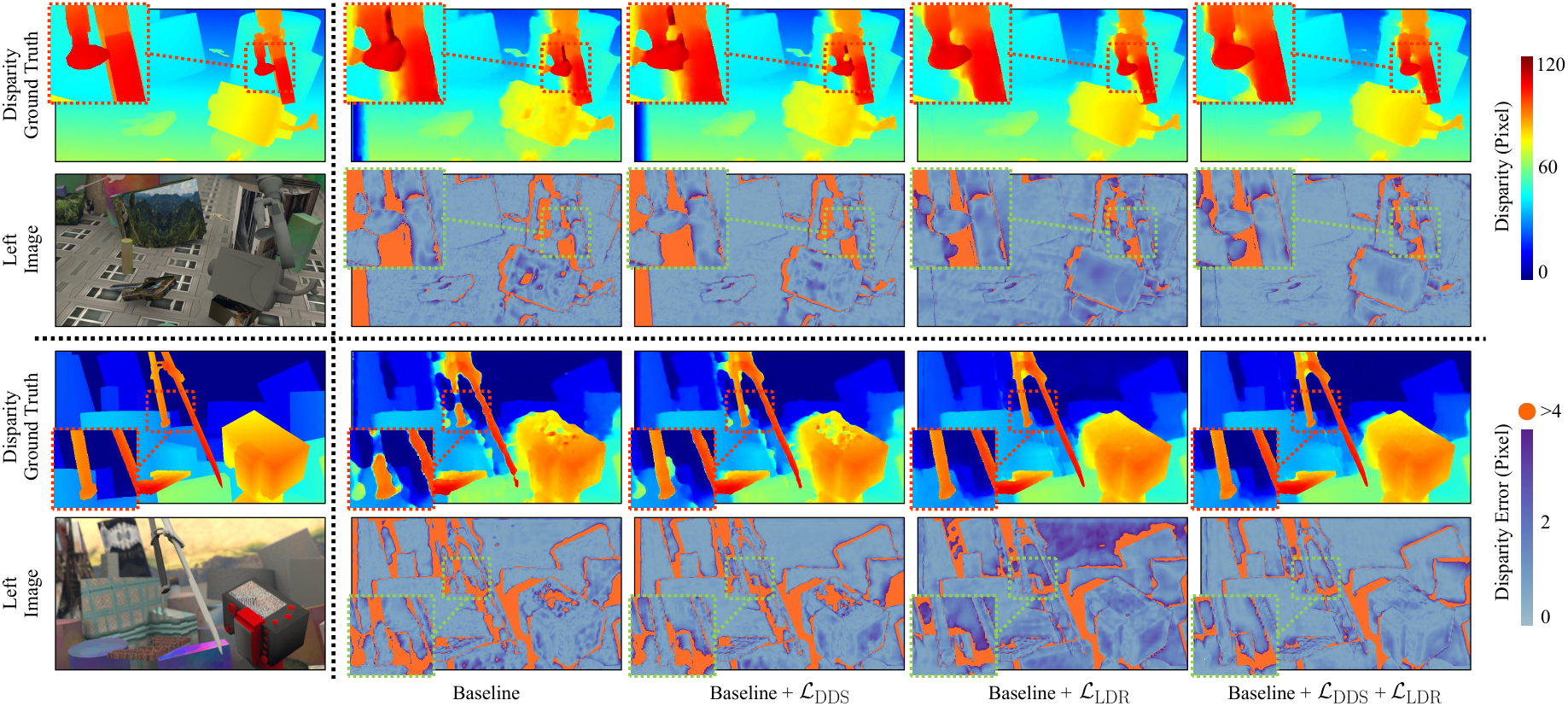}
		\caption{Examples of disparity estimation results with or without our proposed $\mathcal{L}_{\text{DDS}}$ and $\mathcal{L}_{\text{LDR}}$, where detailed regions are shown with dashed boxes.}
		\label{fig.abla}
	\end{center}
\end{figure*}

\begin{figure*}[t!]
	\begin{center}
		\includegraphics[width=0.99\textwidth]{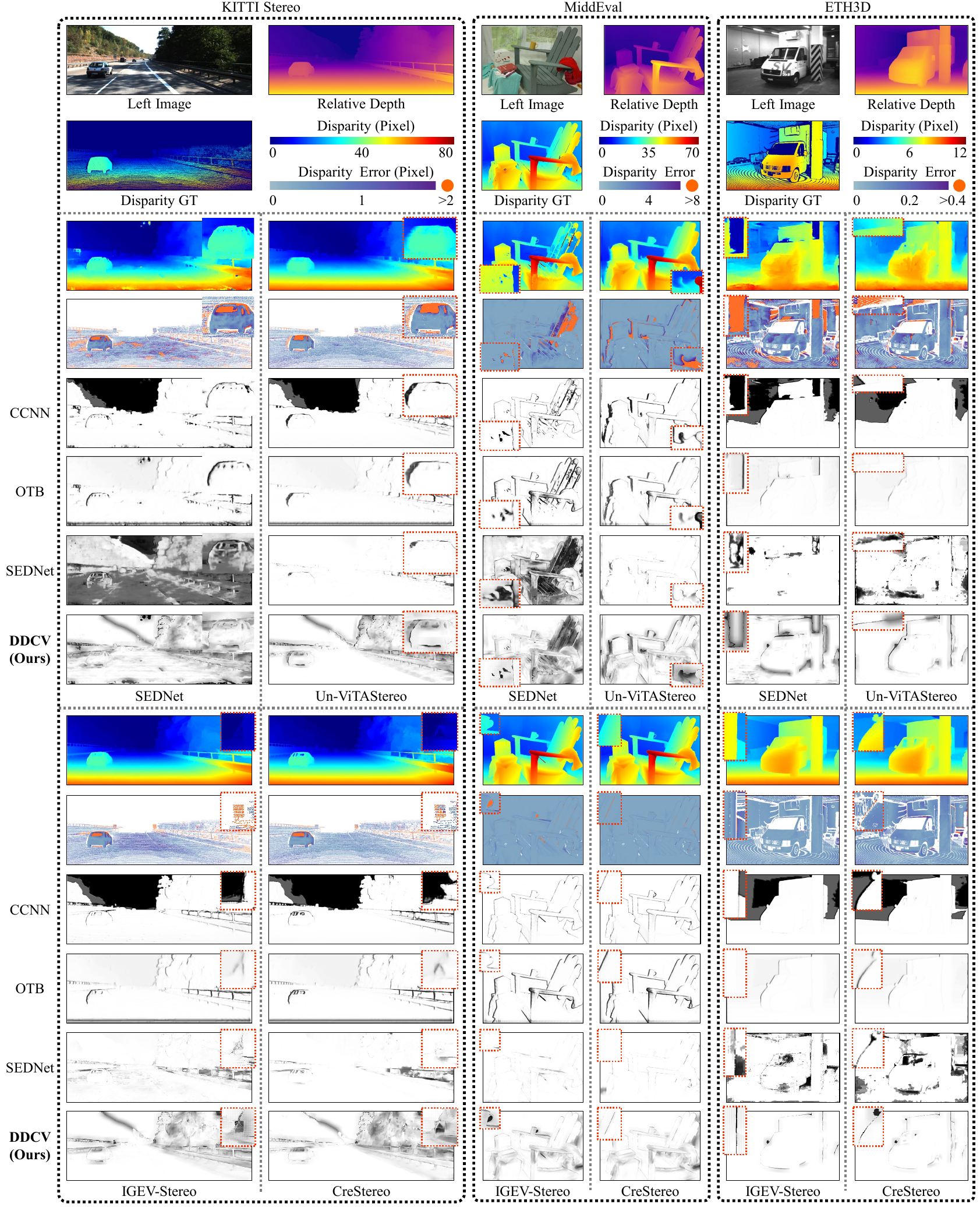}
		\caption{Qualitative experimental results of CCNN \cite{poggi2016learning}, OTB \cite{poggi2020self}, SEDNet \cite{chen2023learning}, and our proposed DDCV on the KITTI Stereo, Middlebury, and ETH3D datasets, where significantly improved regions are shown with dashed boxes.}
		\label{fig.conf}
	\end{center}
\end{figure*}

\subsection{Ablation Study}
\label{sec.abla}

We first investigate the optimal configuration for the top-$k$ operation in $\mathcal{L}_{\text{LDR}}$, referring to the number of reference points for each disparity. As shown in Fig. \ref{fig.topk}, increasing $k$ from zero initially leads to a marked decrease in EPE. This improvement is attributed to the enhanced availability of 3D geometric information provided by a greater number of reference points, which facilitates more accurate localization of the target disparity. However, an excessively large $k$ leads to a sustained increase in EPE. We attribute this degradation to the incorporation of low-confidence disparities into the reference points, which introduces noise into the computation of $\mathcal{L}_{\text{LDR}}$. Therefore, we set $k = 8$, at which the EPE witnesses the first increase, in all subsequent experiments. 

Another ablation study is conducted to demonstrate the superiority of our proposed $\mathcal{L}_{\text{LDR}}$ and $\mathcal{L}_{\text{DDS}}$ over other loss functions related to depth ranking and disparity smoothness, respectively, including the CDR loss $\mathcal{L}_{\text{CDR}}$ proposed in SC-DepthV3 \cite{sun2023sc}, the image-based disparity smoothness loss $\mathcal{L}_{\text{DS}}^{I}$, and the depth-based disparity smoothness loss $\mathcal{L}_{\text{DS}}^{\widetilde{D}}$. The baseline setup includes the photometric loss and left-right consistency loss. As presented in Table \ref{tab.abla}, individually deploying either the CDR loss or LDR loss leads to improved stereo matching accuracy. Specifically, the LDR loss reduces the EPE by approximately 10.7\% compared to the CDR loss. Additionally, the incorporation of both $\mathcal{L}_{\text{CDR}}$ and $\mathcal{L}_{\text{LDR}}$ yields similar results to those of using $\mathcal{L}_{\text{LDR}}$ separately, which signifies the ability of $\mathcal{L}_{\text{LDR}}$ to fully substitute for $\mathcal{L}_{\text{CDR}}$. In terms of the disparity smoothness, our proposed $\mathcal{L}_{\text{DDS}}$ reduces the EPE by 4.5\% and 2.1\% compared with $\mathcal{L}_{\text{DS}}^{I}$ and $\mathcal{L}_{\text{DS}}^{\widetilde{D}}$, respectively. In general, $\mathcal{L}_{\text{LDR}}$ significantly enhances the stereo matching accuracy of ViTAStereo, while $\mathcal{L}_{\text{DDS}}$ provides additional performance gains.

We further present qualitative results of ViTAStereo trained using $\mathcal{L}_{\text{LDR}}$ and $\mathcal{L}_{\text{DDS}}$ in Fig. \ref{fig.abla}. It can be observed that $\mathcal{L}_{\text{DDS}}$ improves the disparity estimation accuracy in both disparity discontinuities and locally smooth regions, aligning well with our intention in Sect. \ref{sec.DDS}. By contrast, $\mathcal{L}_{\text{LDR}}$ exhibits effectiveness in addressing stereo matching ambiguities, including texture-less regions and even the occluded region on the left side of the left image.

\begin{table*}[t!]
    \settablefont
    \caption{Comparisons with SoTA disparity confidence estimation methods. Disparity map inputs are generated by SEDNet \cite{chen2023learning}, our Un-ViTAStereo, IGEVStereo \cite{xu2023iterative} and CreStereo \cite{li2022practical}.}
    \centering
    \label{tab.conf}
    \setlength{\tabcolsep}{1.3mm}{
	\begin{tabular}
		{lcccccccc}
		\toprule[1.5pt]
		\multirow{2}{*}{Dataset} & \multicolumn{2}{c}{Input Disparity Map} & \multicolumn{6}{c}{AUC} \\
		\cmidrule(r){2-3}
		\cmidrule(r){4-9}
		\specialrule{0em}{-1pt}{-1pt}
		& Method & EPE (pixel) & CCNN \cite{poggi2016learning} & LGC \cite{tosi2018beyond} & OTB \cite{poggi2020self} & SEDNet \cite{chen2023learning} & \textbf{DDCV (ours)}&  Optimal \\ 
		\midrule[1pt]
            \multirow{5}{*}{KITTI} & SEDNet \cite{chen2023learning} & 1.16 & 72.5 & 67.6 & 73.1 & 66.9 & \textbf{62.1} & 26.4 \\
    	\cmidrule(r){2-9}
            & Un-ViTAStereo \cite{liu2024playing} & 1.02 & 64.2 & 58.4 & 74.0 & 55.9 & \textbf{54.4} & 25.1\\
    	\cmidrule(r){2-9}
            & IGEVStereo \cite{xu2023iterative} & 1.01 & 75.0 & 70.3 & 79.5 & \textbf{66.8} & 69.0 & 30.1 \\
    	\cmidrule(r){2-9}
            & CreStereo \cite{li2022practical} & 0.92 & 56.2 & 59.2 & 64.5 & \textbf{45.9} & 51.2 & 22.4 \\
    	\midrule[1pt]
            \multirow{5}{*}{Middlebury} & SEDNet \cite{chen2023learning} & 2.24 & 111 & 99.5 & 115 & \textbf{52.1} & 70.8 & 27.2 \\
    	\cmidrule(r){2-9}
            & Un-ViTAStereo \cite{liu2024playing} & 1.96 & 114 & 103 & 113 & 81.9 & \textbf{75.4} & 30.1 \\
    	\cmidrule(r){2-9}
            & IGEVStereo \cite{xu2023iterative} & 1.01 & 76.9 & 65.3 & 70.1 & \textbf{44.6} & 46.1 & 16.7 \\
    	\cmidrule(r){2-9}
            & CreStereo \cite{li2022practical} & 0.64 & 37.1 & 31.9 & 35.7 & 27.8 & \textbf{26.5} & 9.51  \\
    	\midrule[1pt]
            \multirow{5}{*}{ETH3D} & SEDNet \cite{chen2023learning} & 0.78 & 37.3 & 37.1 & 36.8 & 37.6 & \textbf{36.7} & 13.8 \\
    	\cmidrule(r){2-9}
            & Un-ViTAStereo \cite{liu2024playing} & 0.36 & 27.8 & 25.7 & 24.9 & \textbf{17.9} & 21.7 & 9.13 \\
    	\cmidrule(r){2-9}
            & IGEVStereo \cite{xu2023iterative} & 0.09 & 8.38 & 7.50 & 7.92 & \textbf{5.89} & 7.49 & 2.72 \\
    	\cmidrule(r){2-9}
            & CreStereo \cite{li2022practical} & 0.21 & 15.8 & 15.0 & 14.2 & \textbf{12.0} & 13.9 & 5.68 \\
            \bottomrule [1.5pt]   	
	\end{tabular}}
\end{table*}

\begin{table}[t!]
    \settablefont
    \caption{Comparisons with SoTA disparity confidence estimation methods in terms of disparity confidence estimation efficiency.}
    \centering
    \label{tab.runtime}
    \setlength{\tabcolsep}{1.3mm}{
	\begin{tabular}
		{lcccc}
		\toprule[1.5pt]
            Method  & \makebox[0.07\textwidth][c]{CCNN \cite{poggi2016learning}} & \makebox[0.07\textwidth][c]{OTB \cite{poggi2020self}} & \makebox[0.07\textwidth][c]{SEDNet \cite{chen2023learning}} & {\textbf{DDCV (ours)}} \\
		\midrule[1pt]
            time/MP (ms) & 47.2 & 678 & \textbf{6.42} & 9.13 \\
    	\bottomrule [1.5pt]   	
	\end{tabular}}
\end{table}

\begin{figure*}[t!]
	\begin{center}
		\includegraphics[width=0.99\textwidth]{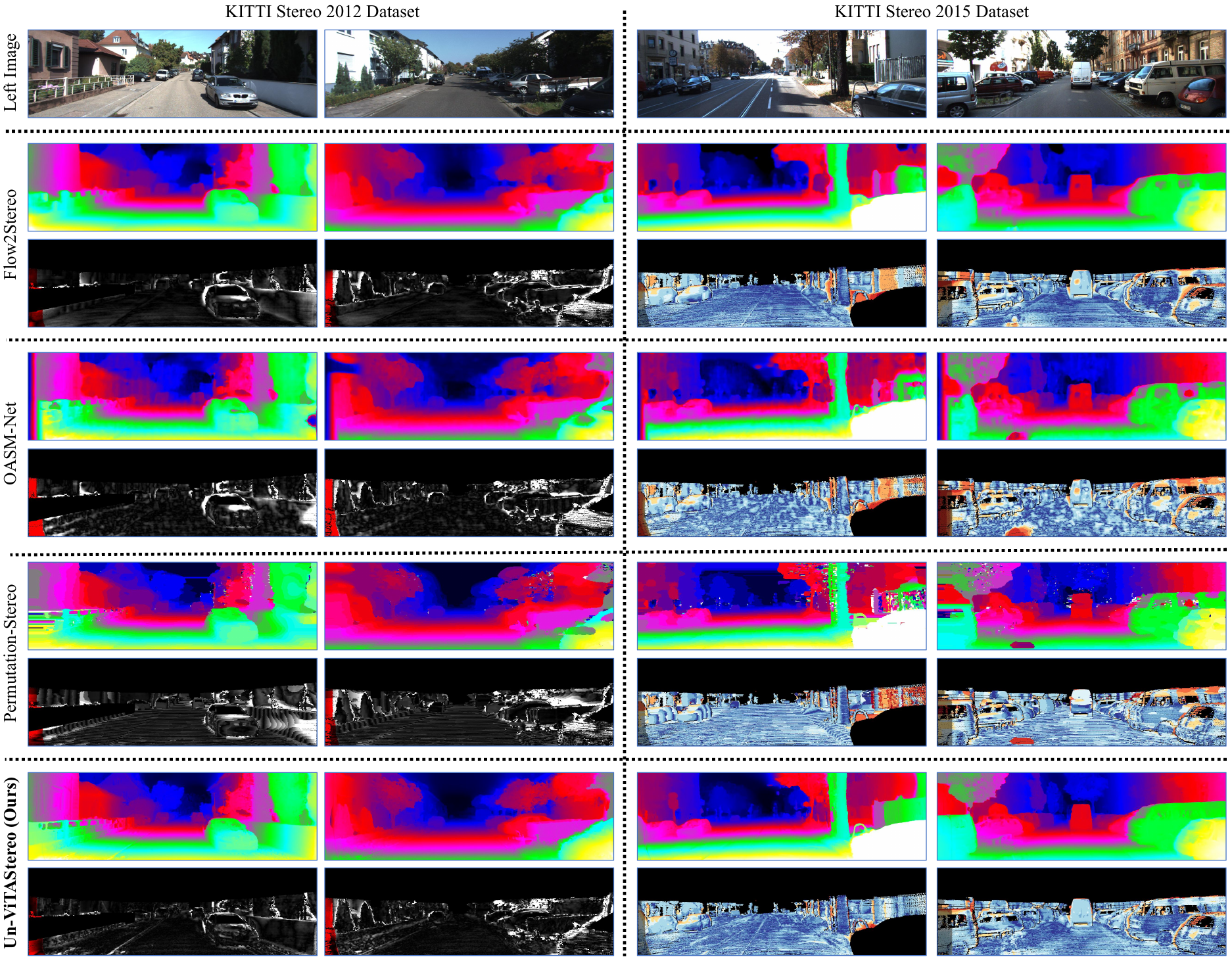}
		\caption{Qualitative experimental results of Flow2Stereo \cite{liu2020flow2stereo}, OASM-Net \cite{li2018occlusion}, Permutation-Stereo \cite{brousseau2022permutation} and this work on the KITTI Stereo datasets \cite{geiger2012we,menze2015object}. The images in the first row of each method represent the estimated disparity maps and images in the second row denote the visualizations of D1 error. We collect these images from the online benchmark suites.} 
		\label{fig.kitti}
	\end{center}
\end{figure*}

\begin{table*}[t!]
    \settablefont
    \caption{Comparisons with SoTA unsupervised stereo matching networks published on the KITTI Stereo benchmark \cite{menze2015object}. ``Noc'' denotes metrics for non-occluded pixels. D1-bg, D1-fg, and D1-all denote D1 for background, foreground, and all pixels, respectively.}
    \centering
    \label{tab.kitti_all}
    \setlength{\tabcolsep}{1.3mm}{
	\begin{tabular}
		{lccccccccccc}
		\toprule[1.5pt]
        \multirow{3}{*}{Network} & \multicolumn{4}{c}{KITTI Stereo 2012} & \multicolumn{6}{c}{KITTI Stereo 2015} & \multirow{3}{*}{Runtime (s)} \\
		\cmidrule(r){2-5}
		\cmidrule(r){6-11}
		\specialrule{0em}{-1pt}{-1pt}
		& \multicolumn{2}{c}{ALL Pixels} & \multicolumn{2}{c}{Noc Pixels} & \multicolumn{3}{c}{All Pixels} & \multicolumn{3}{c}{Noc Pixels} \\
		\cmidrule(r){6-8}
		\cmidrule(r){9-11}
		\cmidrule(r){2-3}
		\cmidrule(r){4-5}
		\specialrule{0em}{-1pt}{-1pt}
		& \makebox[0.058\textwidth][c]{EPE (px)} & \makebox[0.058\textwidth][c]{PEP-3 (\%)} & \makebox[0.058\textwidth][c]{EPE (px)} & \makebox[0.058\textwidth][c]{PEP-3 (\%)} & D1-bg (\%) & D1-fg (\%) & D1-all (\%) & D1-bg (\%) & D1-fg (\%) & D1-all (\%) \\ 
		\midrule[1pt]
        SegStereo \cite{yang2018segstereo} & 2.4 & 9.64 & 1.6 & 7.89 & - & - & 8.79 & - & - & 7.70 & 0.60 \\
        Flow2Stereo \cite{liu2020flow2stereo} & 1.1 & 5.11 & 1.0 & 4.58 & 5.01 & 14.62 & 6.61 & 4.77 & 14.03 & 6.29 & \textbf{0.05} \\
        OASM-Net \cite{li2018occlusion} & 2.0 & 8.60 & 1.3 & 6.39 & 6.89 & 19.42 & 8.98 & 5.44 & 17.30 & 7.39 & 0.73 \\
        PASMnet\_192 \cite{wang2020parallax} & - & - & - & - & 5.41 & 16.36 & 7.23 & 5.02 & 15.16 & 6.69 & 0.50 \\
        OASM-DDS \cite{li2021unsupervised} & 1.2 & 5.69 & 1.1 & 4.81 & 4.46 & 15.76 & 6.51 & - & - & 6.02 & - \\
        Tong et al. \cite{tong2022adaptive} & - & - & - & - & 6.55 & 15.94 & 8.11 & 5.86 & 14.68 & 7.32 & 0.17 \\
        SDCO \cite{cheng2022novel} & - & - & - & - & - & - & - & 5.28 & 17.35 & 7.27 & - \\
        Permutation-Stereo \cite{brousseau2022permutation} & 1.8 & 8.48 & 1.6 & 7.39 & 5.53 &	15.47 &	7.18 & 5.18 & 14.51 & 6.72 & 30.0 \\
        SPSMnet \cite{yang2024self} & - & - & - & - & 5.42 & 12.84 & 6.65 & 4.94 & 12.01 & 6.10 & 0.80 \\ 
        \textbf{Un-ViTAStereo (ours)} & \textbf{0.9} & \textbf{4.16} & \textbf{0.8} & \textbf{3.46} & \textbf{3.58} & \textbf{12.30} & \textbf{5.03} & \textbf{3.28} & \textbf{11.32} & \textbf{4.61} & 0.22 \\ 
        \bottomrule [1.5pt]   	
	\end{tabular}}
\end{table*}

\subsection{Disparity Confidence Estimation}
\label{sec.conf}

We conduct comparative experiments between DDCV and other disparity confidence estimation methods, including CCNN \cite{poggi2016learning}, OTB \cite{poggi2020self}, and SEDNet \cite{chen2023learning}, across various datasets and disparity inputs. All three methods are CNN-based, where CCNN and SEDNet are trained using disparity ground truth, while OTB is trained in an unsupervised manner. The quantitative experimental results are presented in Tables \ref{tab.conf} and \ref{tab.runtime}. It can be observed that DDCV outperforms CCNN, LGC, and OTB across all metrics and achieves comparable AUC scores to those of SEDNet. Specifically, DDCV decreases the AUC score by up to 36.2\%, 28.8\%, and 38.4\% compared with CCNN, LGC, and OTB, respectively. In terms of efficiency, although SEDNet reduces time/MP by 29.6\% compared to DDCV, our method still exhibits real-time performance at the millisecond level. The qualitative experimental results presented in Fig. \ref{fig.conf} show the superiority of our proposed DDCV in identifying erroneous disparities at disparity discontinuities, occlusions, and small-scale objects. In general, our proposed DDCV achieves SoTA disparity confidence estimation accuracy without relying on CNNs and disparity ground truth. We believe that DDCV is able to serve as a strong baseline for future relative depth prior-guided disparity confidence estimation methods.

It is noteworthy that when using disparity maps generated by our unsupervised ViTAStereo (Un-ViTAStereo) as inputs, DDCV achieves lower AUC scores than SEDNet on the KITTI Stereo and Middlebury datasets. The qualitative results on the Middlebury dataset exhibit the superiority of DDCV in handling disparity estimations at occluded areas. The underlying reason is that SEDNet estimates disparity confidence based on the inconsistency between the estimated multi-scale disparities. However, in occluded regions, unsupervised stereo matching networks tend to produce erroneous yet consistent disparities, rendering the multi-scale inconsistency strategy less effective in detecting these unreliable estimates. In contrast, relative depth maps generated by monocular depth estimation networks are able to provide accurate depth ranking information in occluded regions. As a result, our proposed DDCV exhibits robust disparity confidence estimation performance in such regions. Additionally, the absence of ground-truth disparities in occluded regions within the ETH3D dataset accounts for the exceptional case on this dataset. Therefore, our proposed DDCV stands out as a more suitable and effective alternative for integrating unsupervised stereo matching methods.

\subsection{Unsupervised Stereo Matching}
\label{sec.stereo}

Upon submitting our best results to the online KITTI Stereo 2015 and 2012 benchmark suites, we conduct a comparative analysis with other unsupervised stereo matching networks published on these benchmarks, as presented in Table \ref{tab.kitti_all}. Surprisingly, our Un-ViTAStereo outperforms all other SoTA stereo matching networks in terms of all evaluation metrics regarding stereo matching accuracy. Specifically, on the KITTI Stereo 2015 benchmark, Un-ViTAStereo outperforms OASM-DDS \cite{li2021unsupervised}, the second-best network, by up to 22.7\% and 23.4\% in D1 for all pixels and non-occluded pixels, respectively. On the KITTI Stereo 2012 benchmark, Un-ViTAStereo reduces the PEP-3 and EPE by up to 24.5\% and 20\%, respectively, compared with Flow2Stereo \cite{liu2020flow2stereo}. 

The qualitative experimental results, as illustrated in Fig. \ref{fig.kitti}, further demonstrate that our Un-ViTAStereo surpasses other networks in handling stereo matching ambiguities. This advantage is evident in both texture-less regions (illustrated in columns 1 and 2) and repetitive patterns (illustrated in column 3). Additionally, our Un-ViTAStereo achieves higher stereo matching accuracy at disparity discontinuities, such as the vehicle boundaries and background regions (illustrated in column 4). These significant performance gaps demonstrate the effectiveness and efficiency of our proposed $\mathcal{L}_{\text{LDR}}$ and $\mathcal{L}_{\text{DDS}}$ in transferring 3D geometric knowledge from relative depths to a stereo matching network. More visualizations on the KITTI Stereo, Middlebury and ETH3D datasets are provided in our supplementary material at \url{https://mias.group/Un-ViTAStereo} to facilitate further evaluation of the robustness of our proposed Un-ViTAStereo.

\section{Conclusion and Future Work}
\label{sec.conc}

This article introduces a relative depth prior-guided unsupervised stereo matching method, comprising a plug-and-play disparity confidence estimation algorithm and two novel 3D geometric knowledge transfer loss functions. By building quasi-dense correspondences with only reliable disparity estimates, our method improves the knowledge transfer efficiency and accuracy. Our study has yielded several key findings: (1) local coherence consistency between depth priors and estimated disparities is highly indicative of disparity estimation accuracy, and (2) building correspondences with only reliable disparity estimates effectively suppresses noise in the knowledge transfer process. Our method achieves both SoTA stereo matching accuracy on the KITTI Stereo benchmarks, as well as SoTA disparity confidence estimation accuracy among unsupervised methods. 

While the contributions of this study are significant, it is noted that the design of DDCV remains empirical. For instance, we assume pixel correspondence with disparity variation less than 1 pixel as disparity stable. Additionally, the large number of hyper-parameters makes it challenging for our method to achieve optimal performance. Therefore, in future work, we intend to build upon the above-mentioned findings by developing a disparity confidence estimation network that adaptively learns all parameters in an unsupervised manner. Additionally, given that LiDAR can efficiently generate sparse disparity ground truth, enabling the development of real-world datasets including KITTI Stereo, we aim to further explore the potential of LDR loss in supervised stereo matching by selecting pixels with available disparity annotations as reference points.

\normalem
\bibliographystyle{IEEEtran}
\bibliography{refs}

\begin{thebibliography}{10}
\providecommand{\url}[1]{#1}
\csname url@samestyle\endcsname
\providecommand{\newblock}{\relax}
\providecommand{\bibinfo}[2]{#2}
\providecommand{\BIBentrySTDinterwordspacing}{\spaceskip=0pt\relax}
\providecommand{\BIBentryALTinterwordstretchfactor}{4}
\providecommand{\BIBentryALTinterwordspacing}{\spaceskip=\fontdimen2\font plus
\BIBentryALTinterwordstretchfactor\fontdimen3\font minus \fontdimen4\font\relax}
\providecommand{\BIBforeignlanguage}[2]{{%
\expandafter\ifx\csname l@#1\endcsname\relax
\typeout{** WARNING: IEEEtran.bst: No hyphenation pattern has been}%
\typeout{** loaded for the language `#1'. Using the pattern for}%
\typeout{** the default language instead.}%
\else
\language=\csname l@#1\endcsname
\fi
#2}}
\providecommand{\BIBdecl}{\relax}
\BIBdecl

\bibitem{liu2024playing}
C.-W. Liu \emph{et~al.}, ``Playing to vision foundation model's strengths in stereo matching,'' \emph{IEEE Transactions on Intelligent Vehicles}, 2024, {DOI}:10.1109/TIV.2024.3467287.

\bibitem{sun2023sc}
L.~Sun \emph{et~al.}, ``{SC-DepthV3}: Robust self-supervised monocular depth estimation for dynamic scenes,'' \emph{IEEE Transactions on Pattern Analysis and Machine Intelligence}, vol.~46, no.~1, pp. 497--508, 2023.

\bibitem{lu2021resource}
Z.~Lu \emph{et~al.}, ``A resource-efficient pipelined architecture for real-time semi-global stereo matching,'' \emph{IEEE Transactions on Circuits and Systems for Video Technology}, vol.~32, no.~2, pp. 660--673, 2021.

\bibitem{zeng2023deep}
K.~Zeng \emph{et~al.}, ``Deep stereo network with mrf-based cost aggregation,'' \emph{IEEE Transactions on Circuits and Systems for Video Technology}, vol.~34, no.~4, pp. 2426--2438, 2023.

\bibitem{liu2023stereo}
C.-W. Liu \emph{et~al.}, ``Stereo matching: fundamentals, state-of-the-art, and existing challenges,'' in \emph{Autonomous Driving Perception: Fundamentals and Applications}.\hskip 1em plus 0.5em minus 0.4em\relax Springer, 2023, pp. 63--100.

\bibitem{fan2018road}
R.~Fan \emph{et~al.}, ``Road surface {3D} reconstruction based on dense subpixel disparity map estimation,'' \emph{IEEE Transactions on Image Processing}, vol.~27, no.~6, pp. 3025--3035, 2018.

\bibitem{fan2021learning}
R.~Fan \emph{et~al.}, ``Learning collision-free space detection from stereo images: Homography matrix brings better data augmentation,'' \emph{IEEE/ASME Transactions on Mechatronics}, vol.~27, no.~1, pp. 225--233, 2021.

\bibitem{fan2020sne}
R.~Fan \emph{et~al.}, ``{SNE-RoadSeg}: Incorporating surface normal information into semantic segmentation for accurate freespace detection,'' in \emph{European Conference on Computer Vision (ECCV)}.\hskip 1em plus 0.5em minus 0.4em\relax Springer, 2020, pp. 340--356.

\bibitem{liu2023low}
H.~Liu \emph{et~al.}, ``A low-cost and scalable framework to build large-scale localization benchmark for augmented reality,'' \emph{IEEE Transactions on Circuits and Systems for Video Technology}, vol.~34, no.~4, pp. 2274--2288, 2023.

\bibitem{deng2023masic}
X.~Deng \emph{et~al.}, ``{MASIC}: Deep mask stereo image compression,'' \emph{IEEE Transactions on Circuits and Systems for Video Technology}, vol.~33, no.~10, pp. 6026--6040, 2023.

\bibitem{li2023whu}
S.~Li \emph{et~al.}, ``{WHU-Stereo}: A challenging benchmark for stereo matching of high-resolution satellite images,'' \emph{IEEE Transactions on Geoscience and Remote Sensing}, vol.~61, pp. 1--14, 2023.

\bibitem{fan2019pothole}
R.~Fan \emph{et~al.}, ``Pothole detection based on disparity transformation and road surface modeling,'' \emph{IEEE Transactions on Image Processing}, vol.~29, pp. 897--908, 2019.

\bibitem{fan2021graph}
R.~Fan \emph{et~al.}, ``Graph attention layer evolves semantic segmentation for road pothole detection: A benchmark and algorithms,'' \emph{IEEE Transactions on Image Processing}, vol.~30, pp. 8144--8154, 2021.

\bibitem{lee2021high}
Y.~Lee and H.~Kim, ``A high-throughput depth estimation processor for accurate semiglobal stereo matching using pipelined inter-pixel aggregation,'' \emph{IEEE Transactions on Circuits and Systems for Video Technology}, vol.~32, no.~1, pp. 411--422, 2021.

\bibitem{chen2023unambiguous}
Q.~Chen \emph{et~al.}, ``Unambiguous pyramid cost volumes fusion for stereo matching,'' \emph{IEEE Transactions on Circuits and Systems for Video Technology}, vol.~34, no.~10, pp. 9223--9236, 2023.

\bibitem{fan2021rethinking}
R.~Fan \emph{et~al.}, ``Rethinking road surface {3-D} reconstruction and pothole detection: From perspective transformation to disparity map segmentation,'' \emph{IEEE Transactions on Cybernetics}, vol.~52, no.~7, pp. 5799--5808, 2022.

\bibitem{zhang2021farther}
H.~Zhang \emph{et~al.}, ``The farther the better: Balanced stereo matching via depth-based sampling and adaptive feature refinement,'' \emph{IEEE Transactions on Circuits and Systems for Video Technology}, vol.~32, no.~7, pp. 4613--4625, 2021.

\bibitem{li2024inter}
P.~Li \emph{et~al.}, ``Inter-scale similarity guided cost aggregation for stereo matching,'' \emph{IEEE Transactions on Circuits and Systems for Video Technology}, vol.~35, no.~1, pp. 134--147, 2024.

\bibitem{okae2021robust}
J.~Okae \emph{et~al.}, ``Robust scale-aware stereo matching network,'' \emph{IEEE Transactions on Artificial Intelligence}, vol.~3, no.~2, pp. 244--253, 2021.

\bibitem{yao2021decomposition}
C.~Yao \emph{et~al.}, ``A decomposition model for stereo matching,'' in \emph{Proceedings of the Computer Vision and Pattern Recognition Conference (CVPR)}, 2021, pp. 6091--6100.

\bibitem{bartolomei2025stereo}
L.~Bartolomei \emph{et~al.}, ``Stereo anywhere: Robust zero-shot deep stereo matching even where either stereo or mono fail,'' in \emph{Proceedings of the Computer Vision and Pattern Recognition Conference (CVPR)}, 2025, pp. 1013--1027.

\bibitem{cheng2025monster}
J.~Cheng \emph{et~al.}, ``Monster: Marry monodepth to stereo unleashes power,'' in \emph{Proceedings of the IEEE/CVF Conference on Computer Vision and Pattern Recognition (CVPR)}, 2025, pp. 6273--6282.

\bibitem{joung2019unsupervised}
S.~Joung \emph{et~al.}, ``Unsupervised stereo matching using confidential correspondence consistency,'' \emph{IEEE Transactions on Intelligent Transportation Systems}, vol.~21, no.~5, pp. 2190--2203, 2019.

\bibitem{yang2018segstereo}
G.~Yang \emph{et~al.}, ``{SegStereo}: Exploiting semantic information for disparity estimation,'' in \emph{Proceedings of the European Conference on Computer Vision (ECCV)}, 2018, pp. 636--651.

\bibitem{ren2020suw}
H.~Ren \emph{et~al.}, ``{SUW}-learn: Joint supervised, unsupervised, weakly supervised deep learning for monocular depth estimation,'' in \emph{Proceedings of the IEEE/CVF Conference on Computer Vision and Pattern Recognition Workshops (CVPRW)}, 2020, pp. 750--751.

\bibitem{uddin2022unsupervised}
S.~N. Uddin \emph{et~al.}, ``Unsupervised deep event stereo for depth estimation,'' \emph{IEEE Transactions on Circuits and Systems for Video Technology}, vol.~32, no.~11, pp. 7489--7504, 2022.

\bibitem{yang2024self}
X.~Yang \emph{et~al.}, ``Self-supervised learning of {PSMNet} via generative adversarial networks,'' in \emph{International Conference on Intelligent Computing (ICIC)}.\hskip 1em plus 0.5em minus 0.4em\relax Springer, 2024, pp. 469--479.

\bibitem{zhong2017self}
\BIBentryALTinterwordspacing
Y.~Zhong \emph{et~al.}, ``Self-supervised learning for stereo matching with self-improving ability,'' \emph{Computing Research Repository {(CoRR)}}, vol. abs/1709.00930, 2017. [Online]. Available: \url{https://arxiv.org/abs/1709.00930}
\BIBentrySTDinterwordspacing

\bibitem{li2018occlusion}
A.~Li and Z.~Yuan, ``Occlusion aware stereo matching via cooperative unsupervised learning,'' in \emph{Asian Conference on Computer Vision}.\hskip 1em plus 0.5em minus 0.4em\relax Springer, 2018, pp. 197--213.

\bibitem{li2021unsupervised}
A.~Li \emph{et~al.}, ``Unsupervised occlusion-aware stereo matching with directed disparity smoothing,'' \emph{IEEE Transactions on Intelligent Transportation Systems}, vol.~23, no.~7, pp. 7457--7468, 2021.

\bibitem{wang2021co}
H.~Wang \emph{et~al.}, ``Co-{Teaching}: An ark to unsupervised stereo matching,'' in \emph{2021 IEEE International Conference on Image Processing (ICIP)}.\hskip 1em plus 0.5em minus 0.4em\relax IEEE, 2021, pp. 3328--3332.

\bibitem{liu2020flow2stereo}
P.~Liu \emph{et~al.}, ``{Flow2Stereo}: Effective self-supervised learning of optical flow and stereo matching,'' in \emph{Proceedings of the IEEE/CVF conference on Computer Vision and Pattern Recognition (CVPR)}, 2020, pp. 6648--6657.

\bibitem{poggi2021confidence}
M.~Poggi \emph{et~al.}, ``On the confidence of stereo matching in a deep-learning era: a quantitative evaluation,'' \emph{IEEE Transactions on Pattern Analysis and Machine Intelligence}, vol.~44, no.~9, pp. 5293--5313, 2021.

\bibitem{hu2012quantitative}
X.~Hu and P.~Mordohai, ``A quantitative evaluation of confidence measures for stereo vision,'' \emph{IEEE Transactions on Pattern Analysis and Machine Intelligence}, vol.~34, no.~11, pp. 2121--2133, 2012.

\bibitem{poggi2017quantitative}
M.~Poggi \emph{et~al.}, ``Quantitative evaluation of confidence measures in a machine learning world,'' in \emph{Proceedings of the IEEE International Conference on Computer Vision (ICCV)}, 2017, pp. 5228--5237.

\bibitem{poggi2016learningo1}
M.~Poggi and S.~Mattoccia, ``Learning a general-purpose confidence measure based on {O}(1) features and a smarter aggregation strategy for semi global matching,'' in \emph{2016 Fourth International Conference on 3D Vision (3DV)}.\hskip 1em plus 0.5em minus 0.4em\relax IEEE, 2016, pp. 509--518.

\bibitem{poggi2020learning}
M.~Poggi \emph{et~al.}, ``Learning a confidence measure in the disparity domain from {O}(1) features,'' \emph{Computer Vision and Image Understanding}, vol. 193, pp. 2905--2913, 2020.

\bibitem{poggi2020self}
M.~Poggi \emph{et~al.}, ``Self-adapting confidence estimation for stereo,'' in \emph{Proceedings of the European Conference on Computer Vision (ECCV)}.\hskip 1em plus 0.5em minus 0.4em\relax Springer, 2020, pp. 715--733.

\bibitem{poggi2016learning}
M.~Poggi and S.~Mattoccia, ``Learning from scratch a confidence measure,'' in \emph{Proceedings of 27th British Machine Vision Conference (BMVC)}, 2016, pp. 1--13.

\bibitem{tosi2018beyond}
F.~Tosi \emph{et~al.}, ``Beyond local reasoning for stereo confidence estimation with deep learning,'' in \emph{Proceedings of the European Conference on Computer Vision (ECCV)}, 2018, pp. 319--334.

\bibitem{ronneberger2015u}
O.~Ronneberger \emph{et~al.}, ``{U-Net}: Convolutional networks for biomedical image segmentation,'' in \emph{Proceddings of the Medical Image Computing and Cmputer-Assisted Intervention (MICCAI)}.\hskip 1em plus 0.5em minus 0.4em\relax Springer, 2015, pp. 234--241.

\bibitem{chen2023learning}
L.~Chen \emph{et~al.}, ``Learning the distribution of errors in stereo matching for joint disparity and uncertainty estimation,'' in \emph{Proceedings of the IEEE/CVF Conference on Computer Vision and Pattern Recognition (CVPR)}, 2023, pp. 17\,235--17\,244.

\bibitem{tosi2017learning}
F.~Tosi \emph{et~al.}, ``Learning confidence measures in the wild,'' in \emph{Proceedings of 28th British Machine Vision Conference (BMVC)}, 2017, pp. 1--13.

\bibitem{park2018learning}
M.-G. Park and K.-J. Yoon, ``Learning and selecting confidence measures for robust stereo matching,'' \emph{IEEE Transactions on Pattern Analysis and Machine Intelligence}, vol.~41, no.~6, pp. 1397--1411, 2018.

\bibitem{mostegel2016using}
C.~Mostegel \emph{et~al.}, ``Using self-contradiction to learn confidence measures in stereo vision,'' in \emph{Proceedings of the IEEE Conference on Computer Vision and Pattern Recognition (CVPR)}, 2016, pp. 4067--4076.

\bibitem{chen2023vision}
Z.~Chen \emph{et~al.}, ``Vision {Transformer} adapter for dense predictions,'' in \emph{International Conference on Learning Representations (ICLR)}, 2023.

\bibitem{yang2024depthv2}
L.~Yang \emph{et~al.}, ``{Depth Anything V2},'' \emph{Advances in Neural Information Processing Systems (NeurIPS)}, vol.~37, pp. 21\,875--21\,911, 2024.

\bibitem{kirillov2023segment}
A.~Kirillov \emph{et~al.}, ``Segment anything,'' in \emph{Proceedings of the IEEE/CVF International Conference on Computer Vision (ICCV)}, 2023, pp. 4015--4026.

\bibitem{yang2024depth}
L.~Yang \emph{et~al.}, ``Depth {Anything}: Unleashing the power of large-scale unlabeled data,'' in \emph{Proceedings of the IEEE/CVF Conference on Computer Vision and Pattern Recognition (CVPR)}, 2024, pp. 10\,371--10\,381.

\bibitem{roy1999stereo}
S.~Roy, ``Stereo without epipolar lines: A maximum-flow formulation,'' \emph{International Journal of Computer Vision}, vol.~34, no.~2, pp. 147--161, 1999.

\bibitem{liu2025these}
C.-W. Liu \emph{et~al.}, ``These maps are made by propagation: Adapting deep stereo networks to road scenarios with decisive disparity diffusion,'' \emph{IEEE Transactions on Image Processing}, vol.~34, pp. 1516--1528, 2025.

\bibitem{wang2021pvstereo}
H.~Wang \emph{et~al.}, ``{PVStereo}: Pyramid voting module for end-to-end self-supervised stereo matching,'' \emph{IEEE Robotics and Automation Letters}, vol.~6, no.~3, pp. 4353--4360, 2021.

\bibitem{wang2004image}
Z.~Wang \emph{et~al.}, ``Image quality assessment: from error visibility to structural similarity,'' \emph{IEEE Transactions on Image Processing}, vol.~13, no.~4, pp. 600--612, 2004.

\bibitem{mayer2016large}
N.~Mayer \emph{et~al.}, ``A large dataset to train convolutional networks for disparity, optical flow, and scene flow estimation,'' in \emph{Proceedings of the IEEE Conference on Computer Vision and Pattern Recognition (CVPR)}, 2016, pp. 4040--4048.

\bibitem{cabon2020virtual}
\BIBentryALTinterwordspacing
Y.~Cabon \emph{et~al.}, ``Virtual {KITTI} 2,'' \emph{Computing Research Repository {(CoRR)}}, vol. abs/2001.10773, 2020. [Online]. Available: \url{https://arxiv.org/abs/2001.10773}
\BIBentrySTDinterwordspacing

\bibitem{geiger2012we}
A.~Geiger \emph{et~al.}, ``Are we ready for autonomous driving? the {KITTI} vision benchmark suite,'' in \emph{2012 IEEE Conference on Computer Vision and Pattern Recognition (CVPR)}.\hskip 1em plus 0.5em minus 0.4em\relax IEEE, 2012, pp. 3354--3361.

\bibitem{menze2015object}
M.~Menze and A.~Geiger, ``Object scene flow for autonomous vehicles,'' in \emph{Proceedings of the IEEE Conference on Computer Vision and Pattern Recognition (CVPR)}, 2015, pp. 3061--3070.

\bibitem{scharstein2014high}
D.~Scharstein \emph{et~al.}, ``High-resolution stereo datasets with subpixel-accurate ground truth,'' in \emph{Pattern Recognition: 36th German Conference (GCPR)}.\hskip 1em plus 0.5em minus 0.4em\relax Springer, 2014, pp. 31--42.

\bibitem{schops2017multi}
T.~Schops \emph{et~al.}, ``A multi-view stereo benchmark with high-resolution images and multi-camera videos,'' in \emph{Proceedings of the IEEE Conference on Computer Vision and Pattern Recognition (CVPR)}, 2017, pp. 3260--3269.

\bibitem{xu2023iterative}
G.~Xu \emph{et~al.}, ``Iterative geometry encoding volume for stereo matching,'' in \emph{Proceedings of the IEEE/CVF Conference on Computer Vision and Pattern Recognition (CVPR)}, 2023, pp. 21\,919--21\,928.

\bibitem{li2022practical}
J.~Li \emph{et~al.}, ``Practical stereo matching via cascaded recurrent network with adaptive correlation,'' in \emph{Proceedings of the IEEE/CVF Conference on Computer Vision and Pattern Recognition (CVPR)}, 2022, pp. 16\,263--16\,272.

\bibitem{brousseau2022permutation}
P.-A. Brousseau and S.~Roy, ``A permutation model for the self-supervised stereo matching problem,'' in \emph{2022 19th Conference on Robots and Vision (CRV)}.\hskip 1em plus 0.5em minus 0.4em\relax IEEE, 2022, pp. 122--131.

\bibitem{wang2020parallax}
L.~Wang \emph{et~al.}, ``Parallax attention for unsupervised stereo correspondence learning,'' \emph{IEEE Transactions on Pattern Analysis and Machine Intelligence}, vol.~44, no.~4, pp. 2108--2125, 2020.

\bibitem{tong2022adaptive}
K.~W. Tong \emph{et~al.}, ``Adaptive cost volume representation for unsupervised high-resolution stereo matching,'' \emph{IEEE Transactions on Intelligent Vehicles}, vol.~8, no.~1, pp. 912--922, 2022.

\bibitem{cheng2022novel}
X.~Cheng \emph{et~al.}, ``A novel cell structure-based disparity estimation for unsupervised stereo matching,'' \emph{IET Image Processing}, vol.~16, no.~6, pp. 1678--1693, 2022.

\end{thebibliography}

\begin{IEEEbiography}[{\includegraphics[width=1in,clip]{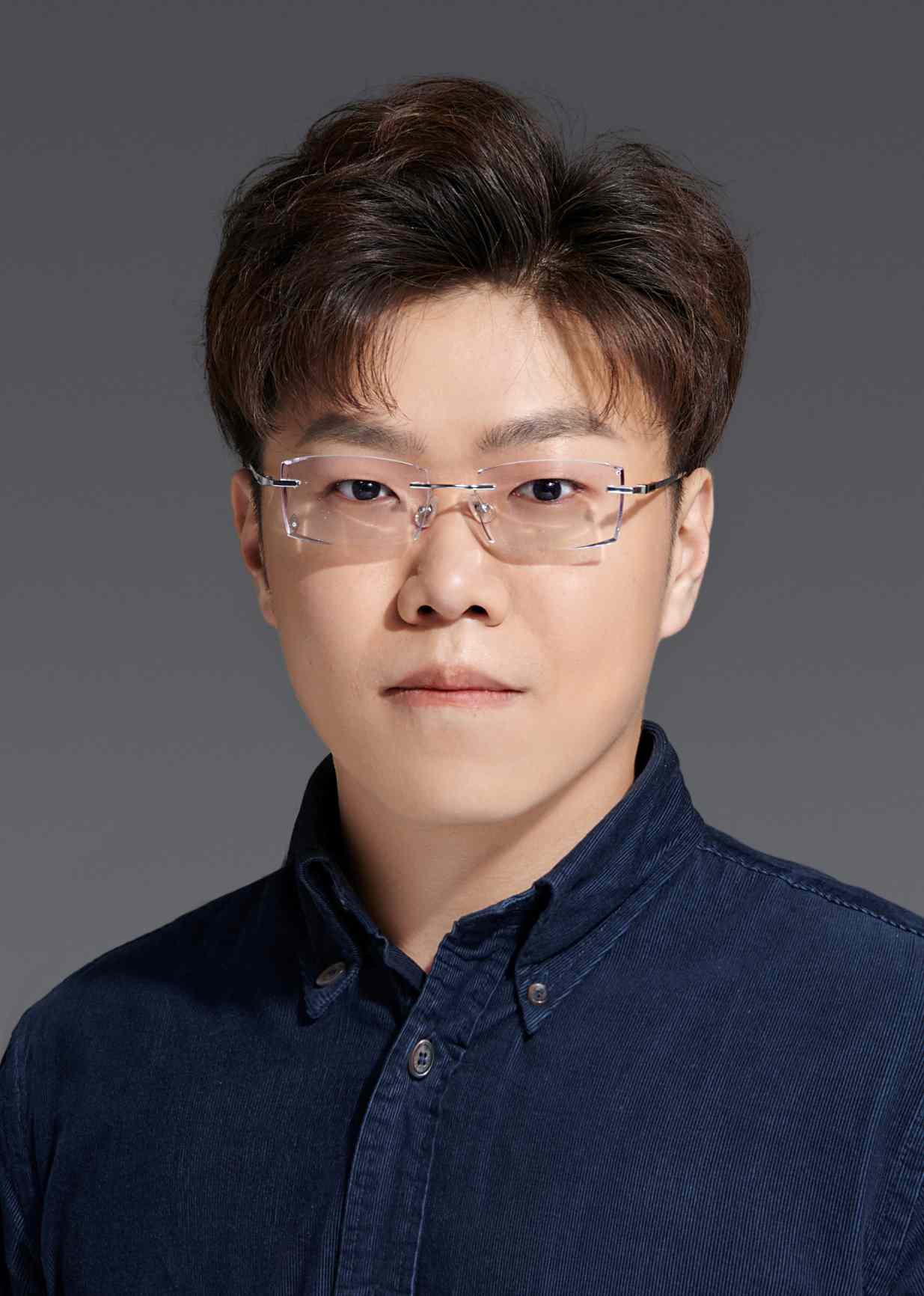}}]{Chuang-Wei Liu} received his B.E. degree in automation from Tongji University in 2020. He is currently pursuing his Ph.D. degree, supervised by Prof. Rui Fan, with the Machine Intelligence and Autonomous Systems (MIAS) Group in the Robotics and Artificial Intelligence Laboratory (RAIL) at Tongji University. His research interests include computer stereo vision, especially for unsupervised approaches, and long-term learning. He is currently a student member of IEEE.
\end{IEEEbiography}

\begin{IEEEbiography}[{\includegraphics[width=1in,clip]{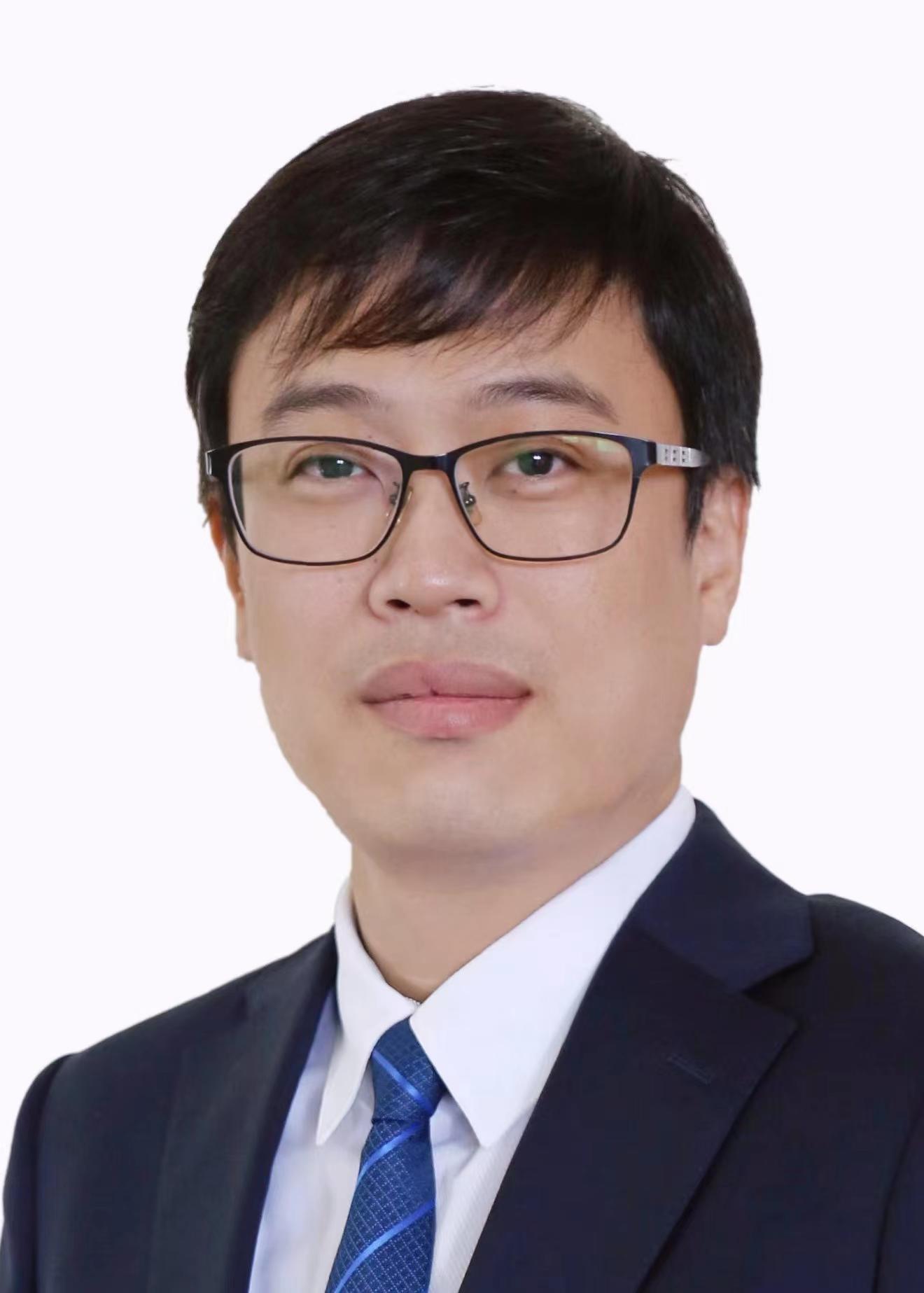}}]{Mingjian Sun}(Member, IEEE) was born in Weifang, Shandong Province, China, in 1980. He received his B.S. degree in Automation in 2003, M.S. degree in Fluid Machinery and Engineering in 2005, and Ph.D. degree in Control Science and Engineering in 2011, all from the Harbin Institute of Technology, China. He became a member of the IEEE in 2016. Professor Sun is currently a Ph.D. supervisor at the Harbin Institute of Technology. His research interests include medical ultrasound and photoacoustic imaging, AI-based medical image analysis, and autonomous control and applications of unmanned systems.
\end{IEEEbiography}

\begin{IEEEbiography}[{\includegraphics[width=1in,clip]{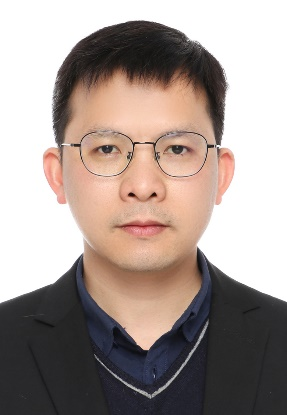}}]{Cairong Zhao} is a Professor at the College of Electronic and Information Engineering, Tongji University. He received his Ph.D. from Nanjing University of Science and Technology in 2011, his M.S. from the Changchun Institute of Optics, Fine Mechanics and Physics, Chinese Academy of Sciences in 2006, and his B.S. from Jilin University in 2003. His research focuses on visual and intelligent learning, with particular interests in computer vision, pattern recognition, and visual surveillance. Professor Zhao has published over 50 papers in leading international conferences and journals, including CVPR, ICCV, ICML, ICLR, AAAI, TPAMI, IJCV, TIP, and TIFS. He currently serves as Chair of the Computer Vision Special Committee of the Shanghai Computer Society. He is also an active reviewer for more than ten top-tier AI-related journals and conferences, including TPAMI, TIP, CVPR, ICCV, NeurIPS, ICML, and AAAI.
\end{IEEEbiography}

\begin{IEEEbiography}[{\includegraphics[width=1in,clip,keepaspectratio]{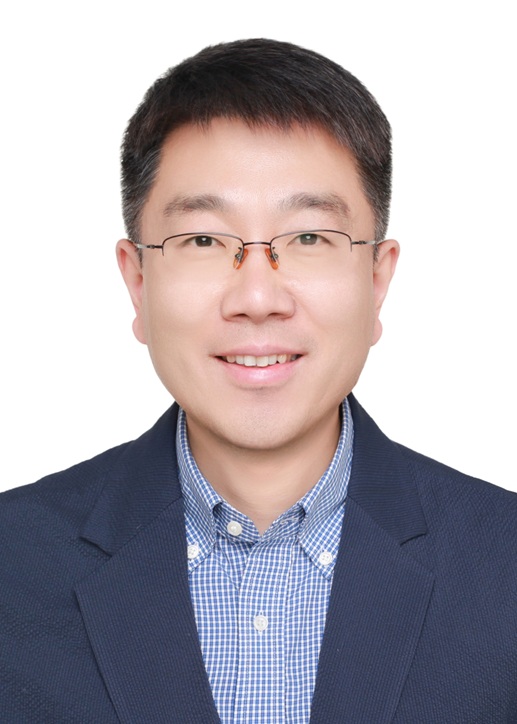}}]{Hanli Wang}(Senior Member) received the B.E. and M.E. degrees in Electrical Engineering from Zhejiang University, Hangzhou, China, in 2001 and 2004, respectively, and the Ph.D. degree in Computer Science from City University of Hong Kong, Hong Kong, in 2007. From 2007 to 2008, he was a Research Fellow in the Department of Computer Science at City University of Hong Kong and concurrently a Visiting Scholar at Stanford University, Palo Alto, CA, USA. From 2008 to 2009, he worked as a Research Engineer at Precoad, Inc., Menlo Park, CA, USA. He was an Alexander von Humboldt Research Fellow at the University of Hagen, Germany, from 2009 to 2010. Since 2010, Dr. Wang has been a Professor with the College of Electronic and Information Engineering at Tongji University, Shanghai, China. His research interests include computer vision, multimedia signal processing, and machine learning.
\end{IEEEbiography}

\begin{IEEEbiography}[{\includegraphics[width=1in,clip]{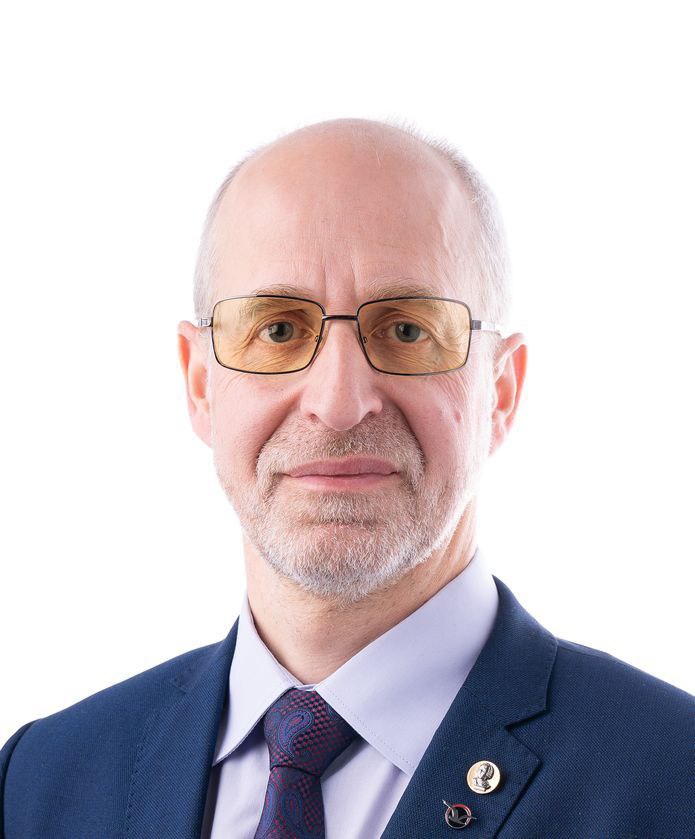}}]{Alexander Dvorkovich} (Member, IEEE) received his M.Sc. degree in Applied Physics and Mathematics from the Moscow Institute of Physics and Technology (MIPT) in 1990, his Ph.D. degree in Chemical Physics (supervisor: Prof. Arkadiy D. Margolin) from MIPT in 1993, and his Doctor of Engineering Science degree in Radio Engineering from Bauman Moscow State Technical University (BMSTU) in 2007. In 2016, he was elected a Corresponding Member of the Russian Academy of Sciences (RAS). He has been working in the field of telecommunications since 1998, initially as a senior researcher at the Radio Research and Development Institute (Moscow, Russia). In 2007, he became Head of the Scientific Research Laboratory at the General Radio Frequency Center (Moscow), and since 2015, he has served as Head of the Multimedia Systems and Technology Laboratory at MIPT. He is currently also Head of the Multimedia Technology and Telecom Department at MIPT and a professor in the Radiotechnical Devices and Antenna Systems Department at the Moscow Power Engineering Institute (National Research University). 
\end{IEEEbiography}

\begin{IEEEbiography}[{\includegraphics[width=1in,clip]{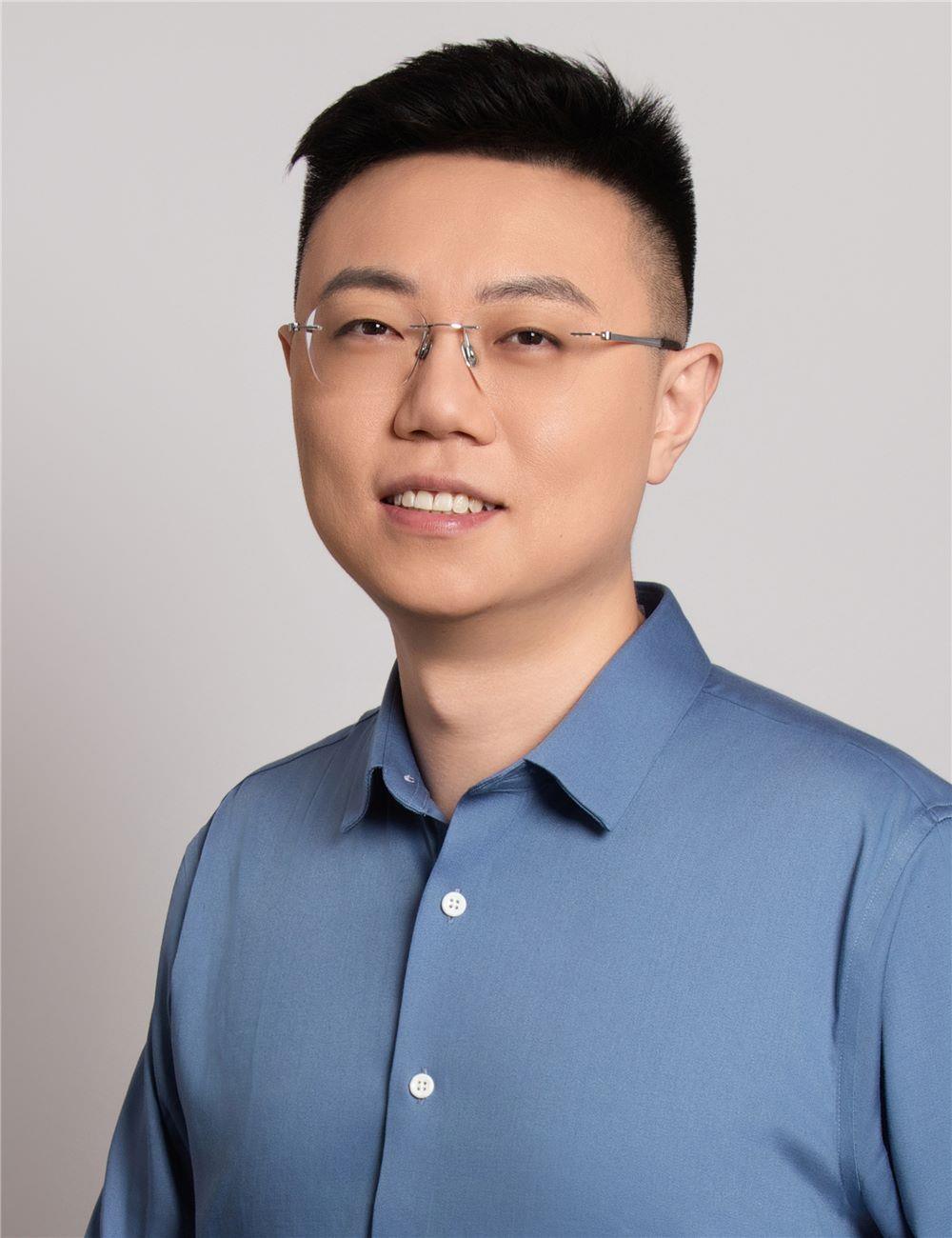}}]{Rui Fan}(Senior Member, IEEE) received the B.Eng. degree in Automation from the Harbin Institute of Technology in 2015 and the Ph.D. degree in Electrical and Electronic Engineering from the University of Bristol in 2018. He worked as a Research Associate at the Hong Kong University of Science and Technology from 2018 to 2020 and a Postdoctoral Scholar-Employee at the University of California San Diego between 2020 and 2021. He began his faculty career as a Full Research Professor in the College of Electronics \& Information Engineering at Tongji University in 2021. He was promoted to Full Professor in 2022 and attained tenure in 2024, both in the same college and at the Shanghai Research Institute for Intelligent Autonomous Systems. His research interests include computer vision, deep learning, and robotics, with a specific focus on humanoid visual perception under the two-streams hypothesis. Prof. Fan served as an associate editor for ICRA'23/25 and IROS'23/24, an area chair for ICIP'24, and a senior program committee member for AAAI'23/24/25. He organized several impactful workshops and special sessions in conjunction with WACV'21, ICIP'21/22/23, ICCV'21/25, and ECCV'22. He was honored by being included in the Stanford University List of Top 2\% Scientists Worldwide between 2022 and 2024, recognized on the Forbes China List of 100 Outstanding Overseas Returnees in 2023, acknowledged as one of Xiaomi Young Talents in 2023, and awarded the Shanghai Science \& Technology 35 Under 35 honor in 2024 as its youngest recipient.
\end{IEEEbiography}
\end{document}